\pdfoutput=1

\documentclass[11pt]{article}

\usepackage{acl}
\usepackage{times}
\usepackage{latexsym}

\usepackage[T1]{fontenc}

\usepackage[utf8]{inputenc}

\usepackage{inconsolata}
\usepackage{booktabs}
\usepackage{enumerate}
\usepackage{enumitem}
\usepackage{amsmath}
\usepackage{amssymb}
\usepackage{algorithm}
\usepackage{algpseudocode}
\usepackage{multirow}
\usepackage{graphicx}
\usepackage{wrapfig}
\usepackage{threeparttable}
\usepackage{bbm}
\usepackage{pifont}
\usepackage[most]{tcolorbox}  
\usepackage{soul}
\usepackage{xcolor}
\definecolor{lightred}{RGB}{255,200,200}

\usepackage{xspace}
\newcommand{\SystemName}{\textsc{nvAgent}\xspace}
\newcommand{\system}{\textsc{nvAgent}\xspace}
\newcommand{\nlvis}{\textsc{NL2Vis}\xspace}

\usepackage{tcolorbox}
\usepackage{xspace}
\tcbuselibrary{skins, breakable, theorems}
\usepackage{colortbl}
\usepackage{subcaption}

\definecolor{darkgreen}{rgb}{0,0.5,0} 
\definecolor{purple}{rgb}{1,0,1} 
\definecolor{todocolor}{rgb}{0.9,0.1,0.1} 
\definecolor{fixcolor}{rgb}{0.1,0.7,0.3} 
\definecolor{wycolor}{rgb}{0.9,0.1,0.1} 
\definecolor{hycolor}{rgb}{0.7,0.7,0.3} 
\definecolor{zwcolor}{rgb}{1,0,1} 
\newcount\DraftStatus  
\newcommand{\nbc}[3]{\ifnum\DraftStatus=1
	{\colorbox{#3}{\bfseries\sffamily\scriptsize\textcolor{white}{#1}}}
	{\textcolor{#3}{\sf\small$\blacktriangleright$\emph{#2}$\blacktriangleleft$}}
\fi}

\newcommand{\draftnote}[2]{\ifnum\DraftStatus=1
	\marginpar{
		\tiny\raggedright
		\hbadness=10000
		\def\baselinestretch{0.8}
		\textcolor{#1}{\textsf{\hspace{0pt}#2}}}
\fi}

\newtcolorbox{promptbox}[1][]{%
  colback=gray!5!white, colframe=gray!75!black, sharp corners, 
  boxrule=0.5mm, top=10pt, bottom=10pt, left=10pt, right=10pt, breakable, title={#1}
}
\title{\includegraphics[height=1.2em]{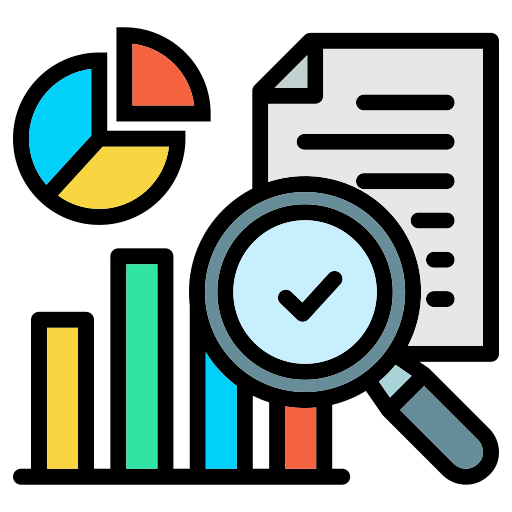}~\SystemName: Automated Data Visualization from Natural Language \\via Collaborative Agent Workflow}

\author{
\textbf{Geliang Ouyang}$^{1}$, \textbf{Jingyao Chen}$^{2}$, \textbf{Zhihe Nie}$^{1}$, \textbf{Yi Gui}$^{1}$, \textbf{Yao Wan}$^{1}$\textsuperscript{\textdagger},\\
\textbf{Hongyu Zhang}$^{3}$, \textbf{Dongping Chen}$^{1}$
\\
$^{1}$Huazhong University of Science and Technology\\
$^{2}$Beijing University of Posts and Telecommunications\ \ \ \ $^{3}$Chongqing University\\
}

\begin{document}
\maketitle
\renewcommand{\thefootnote}{\fnsymbol{footnote}}
\footnotetext[2]{Corresponding anthor (\texttt{wanyao@hust.edu.cn}).}

\begin{abstract}
\textit{Natural Language to Visualization} (\nlvis) seeks to convert natural-language descriptions into visual representations of given tables, empowering users to derive insights from large-scale data.
Recent advancements in \textit{Large Language Models} (LLMs) show promise in automating code generation to transform tabular data into accessible visualizations. However, they often struggle with complex queries that require reasoning across multiple tables. To address this limitation, we propose a collaborative agent workflow, termed \system, for \nlvis. Specifically, \system comprises three agents: \textit{processor} for database processing and context filtering, \textit{composer} for planning visualization generation, and \textit{validator} for code translation and output verification. Comprehensive evaluations on the VisEval benchmark demonstrate that \system consistently surpasses state-of-the-art baselines, achieving 7.88\% and 9.23\% improvements in \textit{single-} and \textit{multi-table} scenarios. Qualitative analyses further highlight that \system maintains nearly a 20\% performance margin over previous methods, underscoring its capacity to produce high-quality visual representations from complex, heterogeneous data sources.\footnote{All datasets and source code are available at: 
\texttt{\url{https://github.com/geliang0114/nvAgent}}. 
A demo video is also provided. We strongly recommend giving a try to visualize multi-table data using chat-style NL instructions.} 
\end{abstract}
\section{Introduction}
\label{sec:intro}
\textit{``Turning data into insight''} has long been a key goal in our increasingly data-rich, information-driven society~\cite{fiorinagoal}. 
To achieve this, \textit{Natural Language to Visualization} (\nlvis) plays a crucial role in transforming natural-language descriptions into visual representations (\emph{e.g.}, charts, plots, and histograms) grounded on tabular data~\cite{sah2024generating}. This approach enables users to interact with data intuitively, facilitating the extraction of patterns and insights from large and complex datasets~\cite{yin2024data,towardsvisualization}.

\begin{figure}[!t]
	\centering
    \includegraphics[width=0.98\linewidth,scale=1.0]
    {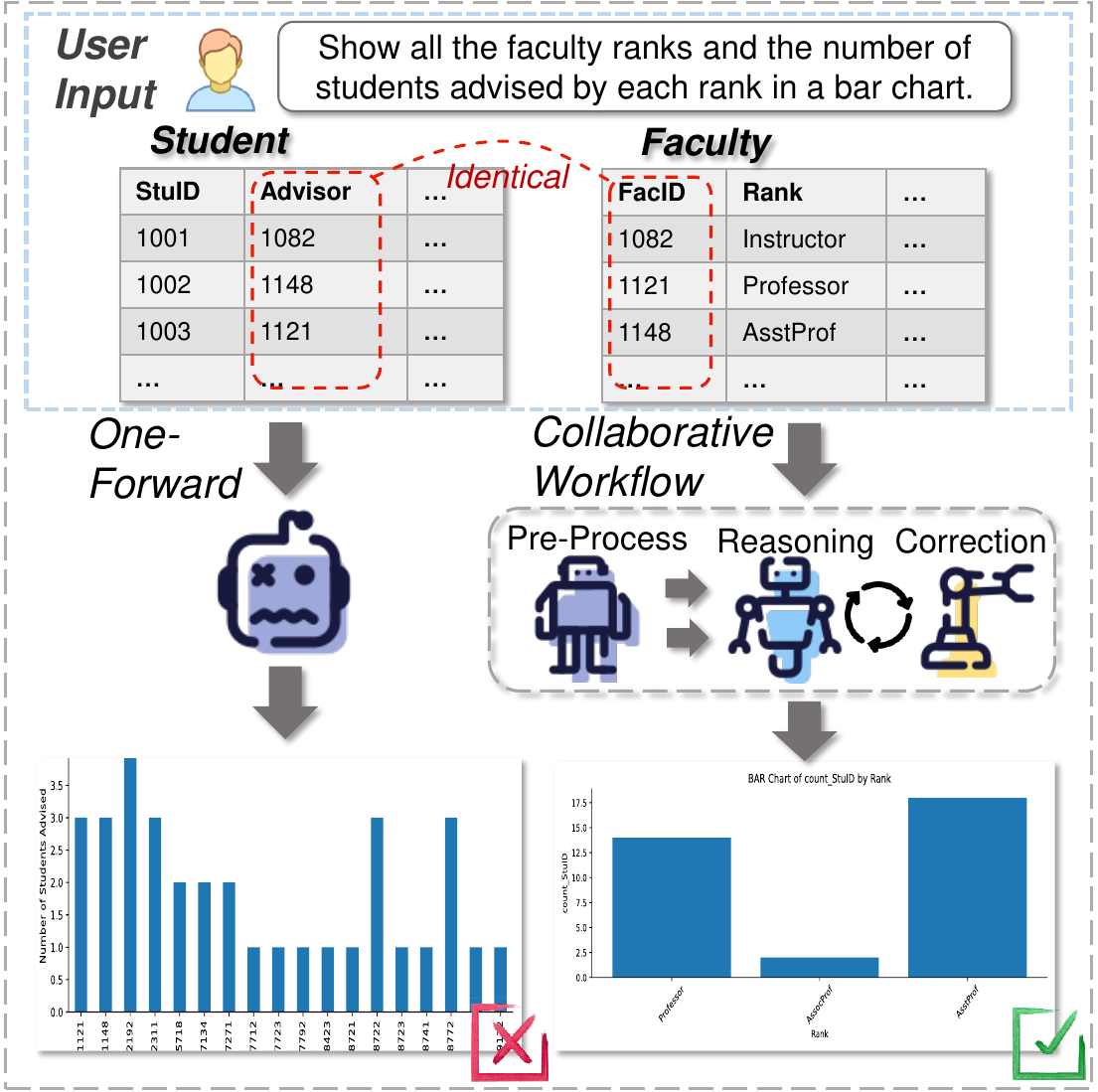}
	\caption{An example to illustrate the \nlvis task. Formerly \textit{``One Forward''} workflow struggled with multi-table queries due to its complex and heterogeneous structure, which could easily cause an error. \system uses a collaborative agent-based workflow for iterative interaction with data and validation to ensure accurate and valid visualization.
    }
    \vspace{-1em}
\label{fig: intro}
\end{figure}

Recently, \textit{Large Language Models} (LLMs) have demonstrated promising performance in \nlvis tasks, excelling in various stages such as data pre-processing~\cite{prompt4vis} and code generation for visualization~\cite{chat2vis}. These models effectively generate readable visualizations for individual datasets or databases~\cite{li2024visualizationgenerationlargelanguage}.
However, existing approaches encounter challenges when processing queries involving multiple tables due to incorrect joins or mis-filtering conditions, leading to visualization errors~\cite{chat2vis,lida,viseval}.  These limitations severely restrict their applicability in real-world scenarios where data is typically distributed across multiple related tables~\cite{khan2024data, lu2024large}.

Figure~\ref{fig: intro} shows an example to illustrate the 
motivation of our study.
Given a natural-language (NL) query such as \textit{``Show all the faculty ranks and the number of students advised by each rank in a bar chart''}, the system must understand that \textit{``faculty''} information corresponds to the column \textit{``Advisor''} and \textit{``FacID''} across two tables. These complex cross-table visualization highlights the challenge between NL queries and databases, requiring a framework that can preprocess metadata, think \textit{``step-by-step''} with plans, and iterative validation to ensure correctness.

These observations inspire our \system, a collaborative agent workflow for \nlvis.
\system follows the \textit{``divide-and-conquer''} paradigm, consisting of three specialized LLM agents: a \textit{processor} agent for database processing and context filtering, a \textit{composer} agent for planning visualization generation, and a \textit{validator} agent for code translation and output verification. 
This collaborative workflow provides a more systematic approach that can effectively handle multi-table scenarios while maintaining visualization accuracy and quality.

To validate the effectiveness of \system, we conducted extensive experiments on the VisEval benchmark~\cite{viseval}, which includes two scenarios: the \textit{single-table} scenario, involving generating visualizations from individual tables, and the \textit{multi-table} scenario, which entails integrating information from multiple tables. The results demonstrate that \system outperforms all baseline methods, achieving a 7.88\% higher pass rate in \textit{single-} and 9.23\% in \textit{multi-table} scenarios compared to the state-of-the-art method. Our ablation study that breakdown every module within \system provide solid evidence of our framework design. Qualitative analyses further highlight that \system maintains 3.64\% and 18.15\% margin in \textit{single-} and \textit{multi-table} over previous frameworks, underscoring its efficacy in producing high-quality visual representations from complex, heterogeneous data sources.

In summary, this paper makes the following key contributions:
    \textbf{(1)} We propose \textbf{\system}, a collaborative agent-based workflow for complex \nlvis tasks, which decomposes the visualization generation process into manageable subtasks.
    \textbf{(2)} Extensive experiments and analysis are performed to validate the effectiveness \textit{divide-and-conquer} strategy of \system for \nlvis.


\section{Problem Formulation}
\label{problem_definition}
A typical workflow of \nlvis tasks involves assembling queries along with tabular data as input, and automatically generating code based on established visualization libraries (\emph{e.g.}, Matplotlib~\cite{barrett2005matplotlib}, Seaborn~\cite{seaborn}) to be executed in a sandboxed environment to obtain the final chart image. 
However, directly generating visualization code often leads to errors due to the complexity of visualization requirements and the semantic gap between natural language and programming constructs.

Following previous works~\cite{nvBench_SIGMOD21,automated}, we introduce \textit{Visualization Query Language} (VQL) as an intermediate representation that bridges natural language queries and visualization code. 
As exemplified below, VQL combines SQL-like syntax for data operations with visualization-specific constructs (\emph{i.e.}, VisType and Binning), making the generation process more controllable and reliable while maintaining simplicity in structure.
\begin{tcolorbox}[sharp corners, colframe=black, colback=white, boxrule=0.5mm, left=1mm, right=1mm, top=1mm, bottom=1mm]
\footnotesize
VisType: \texttt{VISUALIZE \textit{\textcolor{brown}{BAR}}}\\
Data:    \texttt{SELECT \textit{\textcolor{brown}{Date\_Stored, COUNT(Document\_ID)}}} 
\texttt{FROM \textit{\textcolor{brown}{ALL\_Documents}}} \texttt{GROUP BY \textit{\textcolor{brown}{Date\_Stored}}}\\
Binning: \texttt{BIN \textit{\textcolor{brown}{Date\_Stored}} BY \textit{\textcolor{brown}{WEEKDAY}}}
\end{tcolorbox}

Formally, given a natural language query $q$ about a database schema $S$ comprising multiple tables $T$ and columns $C$, the objective of \nlvis is to generate a visualization query $v$ as an intermediate step, which is then translated into a visualization $V$ that accurately represents the data in $S$ to answer the user's query. 

\begin{figure*}[!ht]
	\centering
	\includegraphics[width=0.98\linewidth,scale=1.0]
    {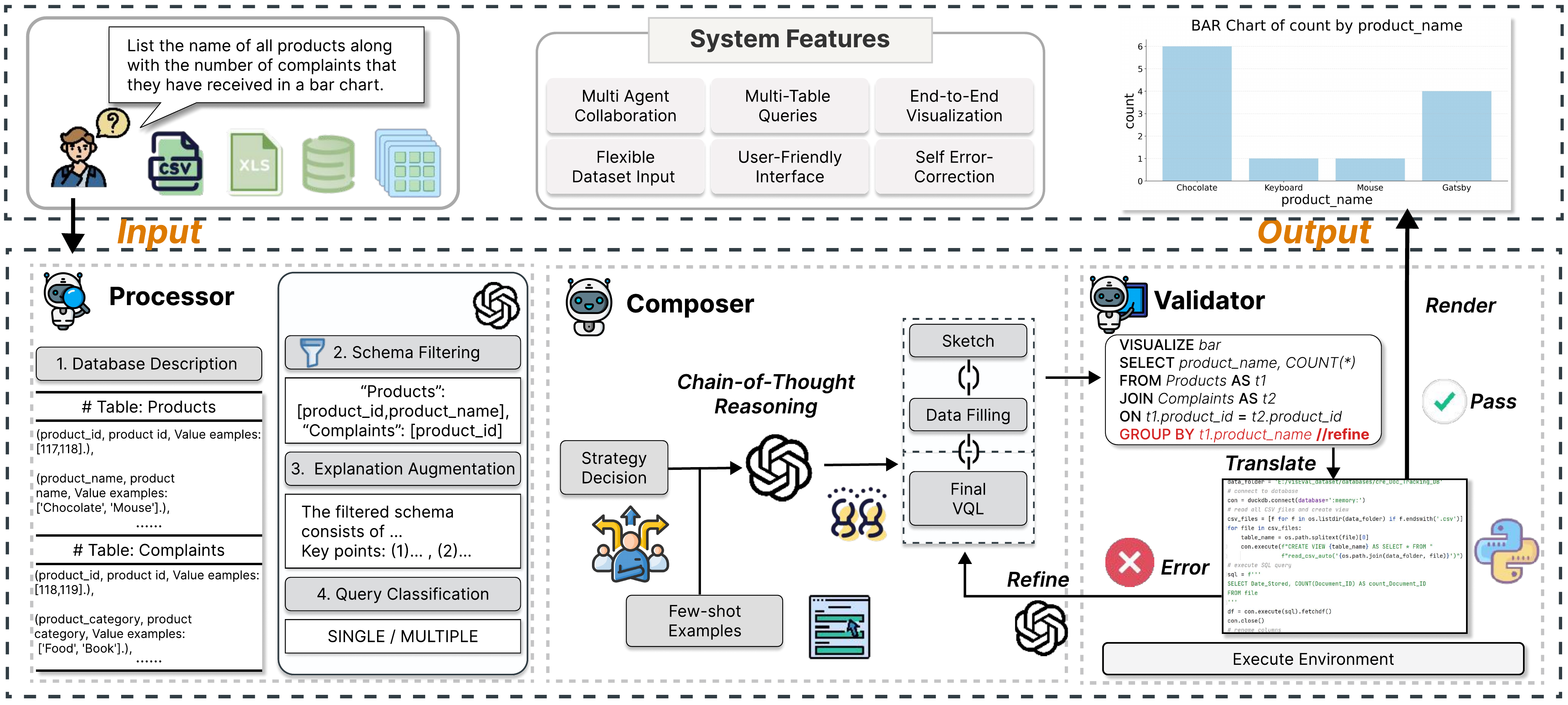}
	\caption{The overall pipeline of \system. We recommend a \textbf{\textit{``Zoom in''}} to view its detailed design:
    \textbf{(1)} The \textit{processor} agent performs schema filtering and context augmentation; \textbf{(2)} The \textit{composer} agent generates structured VQL representations through sketch-and-fill reasoning; \textbf{(3)} The \textit{validator} agent ensures visualization correctness via iterations of execution-guided validation and error-based refinement.}
    \vspace{-1em}
\label{fig: overview}
\end{figure*}

\section{\system: Our Approach}
\label{approach}
\subsection{An Overview}
Figure~\ref{fig: overview} shows an overview of \system, which is composed of three specialized agents: \textit{processor}, \textit{composer}, and \textit{validator}, working collaboratively to transform natural language queries into accurate visualizations.
Starting with a user query $q$ and schema $S$, our approach first leverages the \textit{processor} to filter schema $S'$ and generate additional context including augmented explanation and query complexity classification. The \textit{composer} then generates a VQL query as an intermediate representation through reasoning step by step. Finally, the \textit{validator} ensures correctness via iterative validation and refinement until a valid visualization is produced.


\subsection{Processor Agent} 
\label{processor}

To handle massive data and complex queries effectively, we design a \textit{processor} agent that prepares and enriches input data. 
Specifically, the \textit{processor} agent consists of four steps:

\paragraph{Database Description.}
The \textit{processor} first constructs a comprehensive database description, which includes table and column schemas, with representative value examples. This provides the foundation for LLMs to understand the data structure and relationships. 
For instance, when processing a \textit{``Products''} table, it extracts column details like \textit{``product\_id''}, and \textit{``product\_category''}, along with their value examples (\emph{e.g.}, \textit{``Chocolate''}, \textit{``Book''}).

\paragraph{Schema Filtering.}
Subsequently, building on this foundation, the agent performs schema filtering to identify and extract tables and columns relevant to the user query (\emph{e.g.}, filtering out unrelated columns like \textit{``product\_category''}), effectively reducing noise and preventing information overload.

\paragraph{Explanation Augmentation.}
To enable more accurate query interpretation, inspired by the self-augmented strategy~\cite{augment}, the \textit{processor} generates augmented explanations for the filtered schema like \textit{``Key points: (1) product\_id in the table Products serves as a foreign key linking to the table Complaints''}. These explanations bring insights that provide additional context about table relationships and column semantics. 

\paragraph{Query Classification.}
Finally, the agent classifies query complexity as either \textit{single} or \textit{multiple} based on the number of tables involved and the operations required. This classification guides subsequent agents in choosing appropriate strategies (\emph{e.g.}, the \textit{multiple} scenario requires join operations across tables or complex aggregations).

By providing a focused, well-explained schema and classification, the \textit{processor} agent establishes a strong foundation of complex data understanding for the subsequent stages in our framework.

\subsection{Composer Agent} 
\label{Composer}

The \textit{composer} agent is designed to bridge the gap between natural language queries and visualization code, generating structured VQL queries through a step-by-step reasoning approach. 

\paragraph{Strategy Decision.}
Based on the query classification from the \textit{processor} agent, different strategies are adopted to plan the visualization generation. For example, \textit{single} queries focus on basic aggregations, while \textit{multiple} scenarios require more complex join operations.

\paragraph{Chain-of-Thought Reasoning.}
During the generation stage, the \textit{composer} agent employs a chain-of-thought~\cite{wei2022chain} approach to break down the visualization process into manageable steps. This approach is complemented by providing few-shot examples for In-Context Learning, enhancing the model's adaptability to diverse query types.

\paragraph{Sketch-and-Fill Process.}
The reasoning process follows the \textit{``sketch-and-fill''} paradigm and is structured into three steps, including sketch construction, data components filling, and final VQL composition (prompt shown in Appendix~\ref{prompt_details}).

Taking the query \textit{``List the name of all products along with the number of complaints that they have received in a bar chart.''} (shown in Figure~\ref{fig: overview}) as an example, the \textit{composer} initially determines the specific elements (\emph{i.e.}, visualization type \textit{``Bar''}) and constructs a VQL sketch (\emph{e.g.}, \texttt{``Visualize bar SELECT \_, COUNT(\_) FROM \_ JOIN \_ ON \_''}). Subsequently, it fills the data components (\emph{e.g.}, the column \textit{``product\_name''} ) into the sketch and then combines them to produce the complete VQL query.

\subsection{Validator Agent} 
\label{validator}

The \textit{validator} agent ensures the accuracy and executability of generated VQL queries through an iterative execution-guided validation and error-based refinement process.

\paragraph{Translation and Execution.} 
When receiving a VQL representation, the \textit{validator} first translates the query into executable Python code using visualization libraries like \textit{``Matplotlib''}. The generated code is then executed in a sandboxed environment, where the agent captures either successful execution results or potential error messages.

\paragraph{Pass or Error.} During the execution phase, the \textit{validator} monitors the return information from the execution environment. If successful, it renders and returns the final visualization; otherwise, if errors occur (\emph{e.g.}, syntax errors, or invalid column names), the agent captures specific error messages and routes them back to the \textit{composer} agent, triggering the refinement process.


As shown in Figure~\ref{fig: overview}(c), when the \textit{validator} translates the VQL query \texttt{``VISUALIZE bar ... ON t1.product\_id = t2.product\_id''} into Python code and executes it, an error message \textit{``missing `GROUP BY' clause''} is encountered. This error is then communicated back to the \textit{composer} agent, which refines the VQL query by adding \texttt{``GROUP BY product\_name''} to ensure proper data aggregation.

\paragraph{Iterative Refinement.}
The \textit{composer} agent iteratively refines its output based on feedback from the \textit{validator} agent until a valid visualization is produced. If any errors are detected during validation, it receives error information and adjusts its output accordingly, ensuring the final VQL query is correct. Notably, we design the system to refine VQL query instead of Python code due to its simpler syntax for better correction.

\section{Experiments and Analysis}

\begin{table*}[!t]
\centering
\vspace{-1em}
\begin{threeparttable}
\setlength{\tabcolsep}{2pt} 
\scalebox{0.88}{
\begin{tabular}{l|ccccc|ccccc}
\toprule[1.5pt]
\multirow{2}{*}{\textbf{Method}}& \multicolumn{5}{c|}{\textbf{Single-Table}} & \multicolumn{5}{c}{\textbf{Multi-Table}} \\
& Invalid($\downarrow$) & Illegal($\downarrow$) & Pass($\uparrow$) & Read.($\uparrow$) & Qual.($\uparrow$) & Invalid($\downarrow$) & Illegal($\downarrow$) & Pass($\uparrow$) & Read.($\uparrow$) & Qual.($\uparrow$) \\
\midrule
\multicolumn{11}{c}{\textbf{\textit{GPT-4o}}} \\
\midrule
CoML4Vis & \textbf{0.67\%} & 24.14\% & 75.17\% & 3.42 & 2.58 & 1.87\% & 26.27\% & 71.84\% & 3.45 & 2.48\\
LIDA &  1.13\% & 21.20\% & 77.66\% & 2.53 & 1.99 & 14.80\% & 83.56\% & 1.62\% & 3.62 & 0.06\\
Chat2Vis &  0.86\% & 21.37\% & 77.75\% & \textbf{3.87} & 3.02 & 38.74\% & 59.84\% & 1.40\% & \textbf{3.76} & 0.05\\
\textbf{\system} & 0.72\% & \textbf{13.63\%} & \textbf{85.63\%} & 3.66 & \textbf{3.13} & \textbf{1.34\%} & \textbf{17.57\%} & \textbf{81.07\%} & 3.61 & \textbf{2.93}\\ 
$\Delta$ & \textcolor{red}{-0.05\%} & \textcolor{green!60!black}{+7.57\%} & \textcolor{green!60!black}{+7.88\%} & \textcolor{red}{-5.42\%} & \textcolor{green!60!black}{+3.64\%} & \textcolor{green!60!black}{+0.53\%} & \textcolor{green!60!black}{+8.70\%} & \textcolor{green!60!black}{+9.23\%} & \textcolor{red}{-3.98\%} & \textcolor{green!60!black}{+18.15\%} \\
\midrule
\multicolumn{11}{c}{\textbf{\textit{GPT-4o-mini}}} \\
\midrule
CoML4Vis & \textbf{0.36\%} & 25.74\% & 73.88\% & 3.33 & 2.47 & 10.01\% & 33.06\% & 56.92\% & 3.24 & 1.86\\
LIDA &  9.09\% & 23.04\% & 67.85\% & 3.10 & 2.12 & 17.61\% & 80.86\% & 1.51\% & 3.10 & 0.04\\
Chat2Vis &  2.14\% & 25.92\% & 71.92\% & \textbf{3.81} & 2.76 & 35.78\% & 61.93\% & 2.27\% & 2.30 & 0.05\\
\textbf{\system} & 1.97\% & \textbf{22.86\%} & \textbf{75.16\%} & 3.67 & \textbf{2.77} & \textbf{8.15\%} & \textbf{25.99\%} & \textbf{65.85\%} & \textbf{3.66} & \textbf{2.42}\\ 
$\Delta$ & \textcolor{red}{-1.61\%} & \textcolor{green!60!black}{+0.18\%} & \textcolor{green!60!black}{+1.28\%} & \textcolor{red}{-3.67\%} & \textcolor{green!60!black}{+0.36\%} & \textcolor{green!60!black}{+1.86\%} & \textcolor{green!60!black}{+7.07\%} & \textcolor{green!60!black}{+8.93\%} & \textcolor{green!60!black}{+12.96\%} & \textcolor{green!60!black}{+30.11\%}\\
\midrule
\multicolumn{11}{c}{\textbf{\textit{GPT-3.5-turbo}}} \\
\midrule
CoML4Vis & 6.17\% & 29.28\% & 64.54\% & 3.33 & 2.18 &  13.92\% & 30.09\% & 55.98\% & 3.37 & 1.93 \\
LIDA & 47.32\% & 15.84\% & 36.83\% & 3.32 & 1.23 & 62.57\% & 36.56\% & 0.86\% & 3.50 & 0.03\\
Chat2Vis & 3.90\% & 28.11\% & 67.98\% & 3.03 & 2.08 & 40.77\% & 57.66\% & 1.55\% & 3.31 & 0.05\\
\textbf{\system} & \textbf{2.98\%} & 20.93\% & \textbf{76.08\%} & \textbf{3.58} & \textbf{2.72} & \textbf{7.18\%} & \textbf{28.51\%} & \textbf{64.29\%} & \textbf{3.61} & \textbf{2.32}\\
$\Delta$ & \textcolor{green!60!black}{+0.92\%} & \textcolor{red}{-5.09\%}\textdagger & \textcolor{green!60!black}{+8.10\%} & \textcolor{green!60!black}{+7.51\%} & \textcolor{green!60!black}{+24.77\%} & \textcolor{green!60!black}{+6.74\%} & \textcolor{green!60!black}{+1.58\%} & \textcolor{green!60!black}{+8.11\%} & \textcolor{green!60!black}{+3.14\%} & \textcolor{green!60!black}{+20.21\%}\\
\bottomrule[1.5pt]
\end{tabular}}
\begin{tablenotes}
    \scriptsize
    \item[*] $\Delta$ represents the percentage improvement or decrease of \system compared to the best-performing baseline for each metric. \\
    For the first three columns, $\Delta$ is calculated using absolute differences, while for the last two columns, it is calculated as the relative change. 
    \item \textdagger:~\system actually performs best, while LIDA has a lower Illegal due to its high Invalid rate.
\end{tablenotes}
\vspace{-2mm}
\caption{Performance of our approach with baselines using different backbone models.}
\label{tab:performance_comparison}
\end{threeparttable}
\vspace{-1em}
\end{table*}

\subsection{Experimental Setup}
\label{setup}
\paragraph{Dataset.}
VisEval \cite{viseval} is a benchmark designed based on nvBench \cite{nvbench} to assess the capabilities of LLMs in the \nlvis task. It consists of 1,150 distinct visualizations (VIS) and 2,524 (NL, VIS) pairs across 146 databases, with accurately labeled ground truths and meta-information detailing feasible visualization options. The dataset is divided into \textit{single-table} scenario and \textit{multi-table} scenario. Moreover, visualizations are classified into four distinct levels of hardness: easy, medium, hard, and extra hard. Cases across different hardness levels can be found in Appendix~\ref{example}.

\paragraph{Baselines.} We conduct our experiments compared with three formerly SOTA baselines\footnote{We try the vanilla baseline similar to the GPT-4o with code interpreter in \url{https://platform.openai.com/docs/assistants/tools/code-interpreter}. Due to the API still in the beta stage and often failing, we do not include it as a baseline.}: Chat2Vis \cite{chat2vis}, which uses prompt engineering to generate visualizations from natural language descriptions; LIDA \cite{lida}, which employs a four-step process for incrementally translating natural language inputs into visualizations; and CoML4Vis \cite{coml}, which applies a few-shot prompt method integrating multiple tables for visualization tasks. 
More details can be found in Appendix~\ref{detailed_baselines}.
We implement our approach and baselines using three different backbone models: GPT-4o \citep{openai_gpt4o_2024}, GPT-4o-mini \citep{openai2024gpt4omini}, and GPT-3.5-turbo \citep{chatgpt3.5}. 

\paragraph{Evaluation Metrics.}
We evaluate the performance using both rule-based and model-based metrics for quantitative and qualitative assessment. 
    \textbf{Invalid Rate} and \textbf{Illegal Rate} represent the percentages of visualizations that fail to render or meet query requirements, respectively. 
    \textbf{Pass Rate} measures the proportion of valid and legal visualizations in the evaluation set.
    \textbf{Readability Score} is the average score ranging from 0 to 5 assigned by MLLM-as-a-Judge \citep{chen2024mllm, ye2024justice} to assess their visual clarity for legal visualization. We assess MLLM-scoring by calculating the similarity of GPT-4o-mini and GPT-4o with human-annotated scores in a subset with 500 samples. 
    Empirically, we select GPT-4o-mini as the vision model for judgment. More details are referred to the Appendix~\ref{detailed_experiment_setups}.
    \textbf{Quality Score} is 0 for invalid or illegal visualizations, otherwise equal to the readability score.



\subsection{Overall Performance}
\label{results}

Table~\ref{tab:performance_comparison} shows the performance across different methods and backbone models. Generally, our proposed method, \system, demonstrates significant improvements over existing approaches across all metrics in both \textit{single-} and \textit{multi-table} scenarios, particularly on pass rate and quality score.
Furthermore, \system achieves an impressive 85.63\% pass rate and a quality score of 3.13 in \textit{single-table} scenarios using GPT-4o, surpassing all baseline methods. In more complex \textit{multi-table} scenarios, \system maintains strong performance, significantly outperforming other approaches. Specifically, using GPT-4o, our method attains an 81.07\% pass rate and a 2.93 quality score for \textit{multi-table} queries, exceeding the previous state-of-the-art by 18.15\%. 
The minimal gap between \textit{single-} and \textit{multi-table} scenarios (85.63\% vs. 81.07\%) underscores \system's consistency and adaptability across varying complexities, a crucial advantage in real-world applications where multi-table queries are common.

\begin{table*}[!t]
\centering
\vspace{-1em}
 \setlength{\tabcolsep}{4pt} 
\vspace{-1em}
\scalebox{0.9}{
\begin{tabular}{l|ccc|ccc|c}
\toprule[1.5pt]
\multirow{2}{*}{\textbf{Method}} & \multicolumn{3}{c|}{\textbf{Single-Table}} & \multicolumn{3}{c|}{\textbf{Multi-Table}} & \textbf{Average} \\
& Invalid & Illegal & Pass & Invalid & Illegal & Pass & Pass Rate\\
\midrule
\multicolumn{8}{c}{\textbf{\textit{GPT-4o}}}\\
\midrule
\textbf{\system}(4-shot) & 0.72\% & 13.63\% & \textbf{85.63\%} & 1.34\% & 17.57\% & 81.07\% & 83.80\% \\
w/o Processor & 0.62\% & 14.27\% & 85.09\% & 1.26\% & 16.42\% & \textbf{82.31\%} & \textbf{83.97\%} \\
w/o Composer & 1.20\% & 74.56\% & 24.22\% & 2.34\% & 74.00\% & 23.64\% & 23.99\% \\
w/o Validator & 5.80\% & 12.22\% & 81.96\% & 7.01\% & 15.95\% & 77.02\% & 79.98\% \\
\midrule
\multicolumn{8}{c}{\textbf{\textit{GPT-3.5-turbo}}}\\
\midrule
\textbf{\system}(4-shot) & 2.98\% & 20.93\% & 76.08\% & 7.18\% & 28.51\% & \textbf{64.29\%} & \textbf{71.35\%} \\
w/o Processor & 3.01\% & 20.15\% & \textbf{76.82\%} & 9.38\% & 31.01\% & 59.60\% & 69.92\% \\ w/o Composer & 18.78\% & 30.97\% & 50.24\% & 25.02\% & 27.92\% & 47.05\% & 48.96\% \\ w/o Validator & 18.04\% & 17.50\% & 64.45\% & 22.64\% & 21.40\% & 55.94\% & 61.04\% \\
\bottomrule[1.5pt]
\end{tabular}%
}

\caption{Ablation results of each agent within \system.}
 \label{tab:ablation_agent}
\end{table*}

\begin{table*}[!t]
\centering

\scalebox{0.9}{
\begin{tabular}{l|ccc|ccc|c}
\toprule[1.5pt]
\multirow{2}{*}{\textbf{Method}} & \multicolumn{3}{c|}{\textbf{Single-Table}} & \multicolumn{3}{c|}{\textbf{Multi-Table}} & \textbf{Average} \\
& Invalid & Illegal & Pass & Invalid & Illegal & Pass & Pass Rate\\
\midrule
\textbf{nvAgent}(4-shot) & 2.98\% & 20.93\% & 76.08\% & 7.18\% & 28.51\% & \textbf{64.29\%} & \textbf{71.35\%} \\
\midrule
w/o schema filtering & 3.36\% & 20.09\% & \textbf{76.53\%} & 12.08\% & 30.14\% & 57.77\% & 69.01\% \\
w/o aug. explanation & 3.23\% & 20.69\% & 76.06\% & 7.10\% & 30.87\% & 62.01\% & 70.44\% \\
w/o complex. classifi. & 4.77\% & 21.42\% & 73.79\% & 7.50\% & 29.80\% & 62.69\% & 69.34\% \\
w/o CoT & 15.81\% & 16.91\% & 67.27\% & 17.73\% & 24.40\% & 57.86\% & 63.50\% \\
w/o ICL & 26.80\% & 24.92\% & 48.27\% & 31.91\% & 28.41\% & 39.66\% & 44.82\% \\
\bottomrule[1.5pt]
\end{tabular}%
}
\vspace{-2mm}
\caption{ Ablation results of each module within \system's agentic workflow. }
\label{tab:ablation_tech}
\vspace{-1em}
\end{table*}

\begin{figure}[!t]
    \centering
    \includegraphics[width=0.96\linewidth]{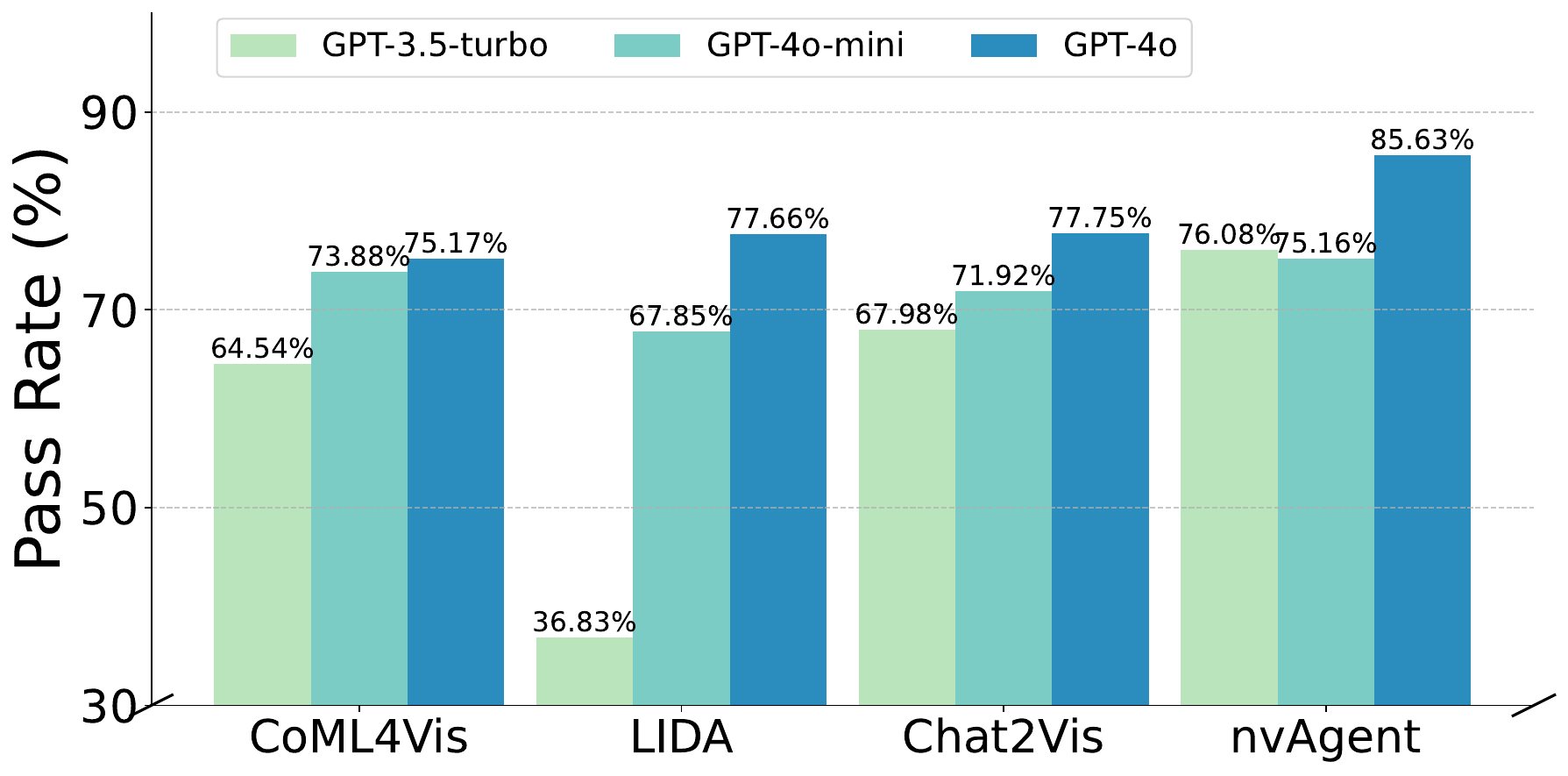}
    \caption{Integrating better LLMs as backbones (\emph{i.e.}, GPT-4o) can bring higher pass rates.
    }
    \vspace{-1em}
    \label{fig:pass_rate}
\end{figure}

\subsection{Effectiveness of Each Agent}
To evaluate the effectiveness of each component in \system, we conducted comprehensive ablation experiments. We perform agent workflow ablation studies with GPT-4o to assess the contributions of each agent, as shown in Table~\ref{tab:ablation_agent}.
From this table, we observe that the \textit{composer} is the most critical component, as its removal leads to significant drops in the overall pass rate—22.39\% with GPT-3.5-turbo and 59.81\% with GPT-4o.
The \textit{validator} also proves vital, as its absence leads to a 3.82\% decrease for GPT-4o and a sharper decrease of 10.31\% using GPT-3.5-turbo, primarily due to increased invalid rate, confirming the effectiveness of the post-processing stage. 

\begin{figure}[!t]
	\centering
	\includegraphics[width=0.9\linewidth,scale=1.0]
    {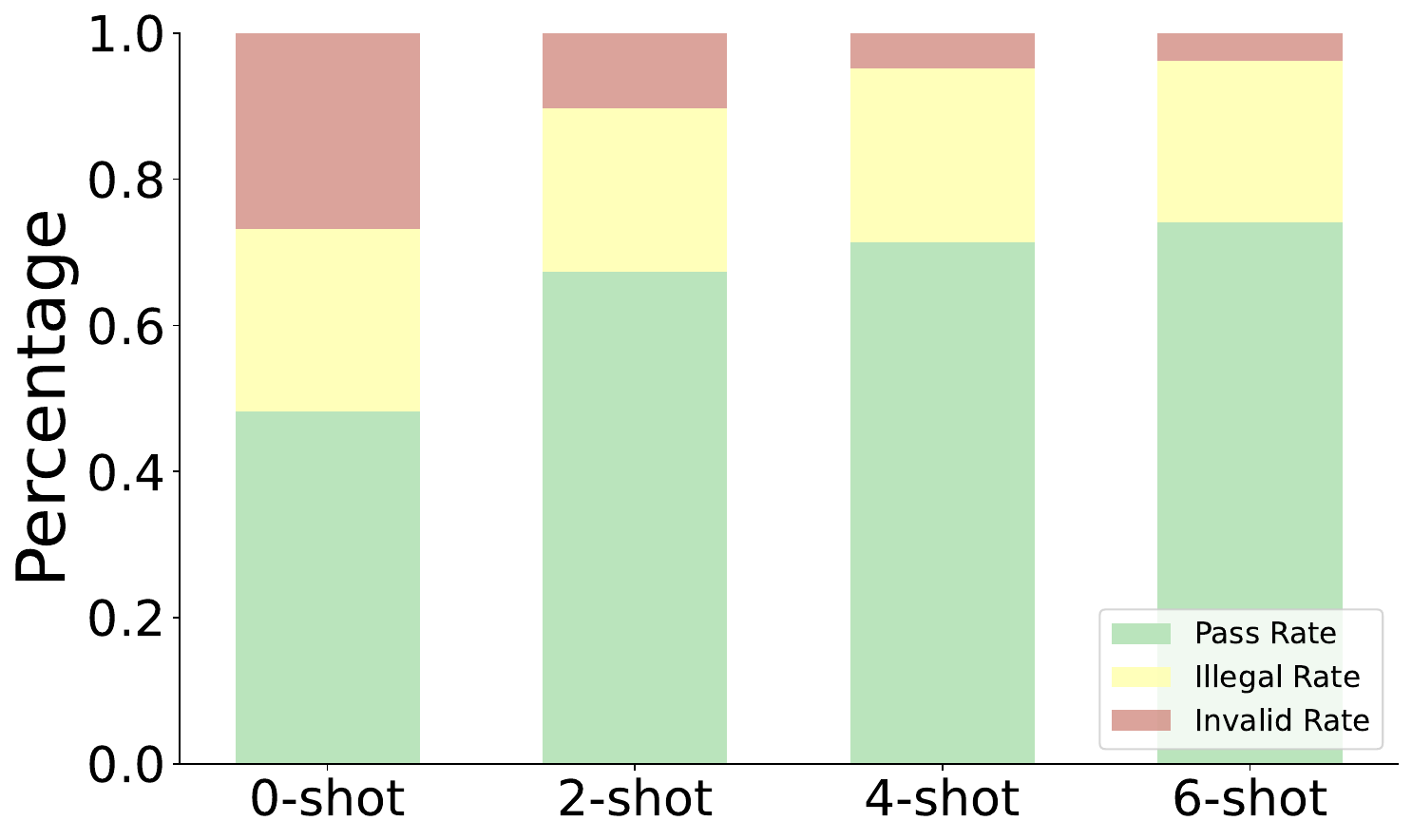}
    \vspace{-2mm}
	\caption{More examples for in-context learning bring higher pass rate, using GPT-3.5-turbo.}
\label{fig: few-shot-ablation}
\vspace{-1em}
\end{figure}

Interestingly, while the \textit{processor}'s removal shows only a slight overall performance decline (1.43\%), its impact varies across scenarios: a marginal improvement in \textit{single-table} cases but a notable decrease (4.69\%) in \textit{multi-table} scenarios. This pattern is particularly pronounced when using GPT-3.5-turbo, highlighting the \textit{processor}'s critical role in handling complex database information. However, more capable models like GPT-4o may occasionally find this additional processing step redundant, as similarly observed in \textit{``The Death of Schema Linking''}~\cite{death}.

\subsection{Impact of LLM Backbones}
Figure~\ref{fig:pass_rate} illustrates the performance of different methods across three backbone LLMs in \textit{single-table} scenarios.
It can be observed that the pass rate positively correlates with the capacity of the backbone LLMs.
However, an intriguing phenomenon was noted: using GPT-4o-mini resulted in a slight decrease in performance compared to GPT-3.5-turbo.
This unexpected outcome suggests potential limitations in GPT-4o-mini's reasoning abilities for this specific task, despite its overall advancements. 
\subsection{Impact of Prompting Techniques}
Further ablation results of individual prompting techniques within each agent using GPT-3.5-turbo are demonstrated in Table~\ref{tab:ablation_tech}. From this table, we observe that all three techniques in \textit{processor} show similar results. However, the schema filtering proves more beneficial for \textit{multi-table} scenarios (6.52\%), while complexity classification benefits \textit{single-table} scenarios (2.29\%). In the \textit{composer} agent, the sharp decrease (26.53\%) upon removal of in-context learning demonstrates the critical role of example-based prompts in task comprehension, and the significant increase in Invalid Rate also highlights the step-by-step VQL generation.
Moreover, as shown in Table~\ref{code_refine}, we conduct an exploration study for \textit{validator} to refine Python code directly and find that the pass rate decreased by 1.01\%, indicating the effectiveness of using VQL for correction.
We also include several exploration experiments in Appendix~\ref{additional_experiment_result}.

\begin{table}[t]
\centering
\setlength{\tabcolsep}{3pt} 
\vspace{-1em}
\scalebox{0.9}{
\begin{tabular}{l|ccc|c}
\toprule[1.5pt]
\textbf{Setting} & \textbf{Invalid} & \textbf{Illegal} & \textbf{Pass} & \textbf{Tokens}\\
\midrule
VQL Refine & 4.66\% & \textbf{23.97\%} & \textbf{71.36\%} & \textbf{1179}\\
Code Refine & \textbf{4.11\%} & 25.51\% & 70.35\% & 1365\\
\bottomrule[1.5pt]
\end{tabular}
}
\caption{Exploration study of Python code refinement. Tokens represent the usage in the refinement stage.}
\label{code_refine}
\end{table}

\begin{table}[!t]
\centering
\small
\renewcommand\arraystretch{1.05}
\setlength{\tabcolsep}{3.5pt}


\begin{tabular}{lcc|cc}
\toprule[1.5pt]
\multirow{2}{*}{\textbf{Method}} & \multicolumn{2}{c|}{\textbf{Single-Table}} & \multicolumn{2}{c}{\textbf{Multi-Table}} \\
\cmidrule(lr){2-3} \cmidrule(lr){4-5}
 & Elo & 95\% CI & Elo & 95\% CI \\
\midrule
\textbf{\system} & \textbf{1538.27} & +2.95/-2.95 & \textbf{1529.86} & +2.83/-2.84 \\
CoML4Vis & 1506.71 & +3.00/-3.00 & 1514.96 & +3.00/-3.00 \\
Chat2Vis & 1496.71 & +3.05/-3.05 & 1499.44 & +3.01/-3.01 \\
LIDA & 1458.31 & +2.85/-2.85 & 1455.74 & +2.94/-2.93 \\
\bottomrule[1.5pt]
\end{tabular}
\caption{Elo rankings on \textit{single-} and \textit{multi-table} test sets. \system scores the highest in both scenarios.}
\label{tab:elo_rankings}
\vspace{-1em}
\end{table}

We carefully design diverse examples including various visualization types (\emph{e.g.}, grouping scatter) and binning operations (\emph{e.g.}, Year, Weekday) for prompting LLM, and Figure~\ref{fig: few-shot-ablation} illustrates the impact of increasing the number of examples in the prompt. The observed improvement in pass rate suggests that the language model effectively leverages knowledge from few-shot prompts.

\subsection{Qualitative Analysis}

\paragraph{ELO Score.}
We adopt the ELO rating system~\cite{elo1978rating}, a widely-used method for calculating relative skill levels, to evaluate model performance.
We conduct this experiment in 1,000 example pairs from \textit{single-} and \textit{multi-table} datasets with equal weights for different models, using human judgments to assess the accuracy of natural language queries. The results in Table~\ref{tab:elo_rankings} show that our \system outperforms other baselines, highlighting its capability to manage complex queries and produce relevant visualizations. Implementation details of the ELO rating framework are in Appendix~\ref{human}.

\paragraph{Case Study.}
Figure~\ref{fig:case_study} presents three cases illustrating NL queries and their visualizations generated by \system and baseline models. The examples showcase \system's superior performance. In the first case, \system correctly orders data by the X-axis, while Chat2Vis and CoML4Vis use the Y-axis. The second case highlights \system's accurate grouping in a stacked bar chart, unlike the baselines. In the third case, involving a \textit{multi-table} query, \system effectively joins tables and groups data for a line chart, whereas Chat2Vis struggles with the structure, and CoML4Vis overlooks the where condition.

\begin{figure}[!t]
    \centering
    \begin{minipage}[b]{\linewidth}
        \centering
        \includegraphics[width=0.76\linewidth]{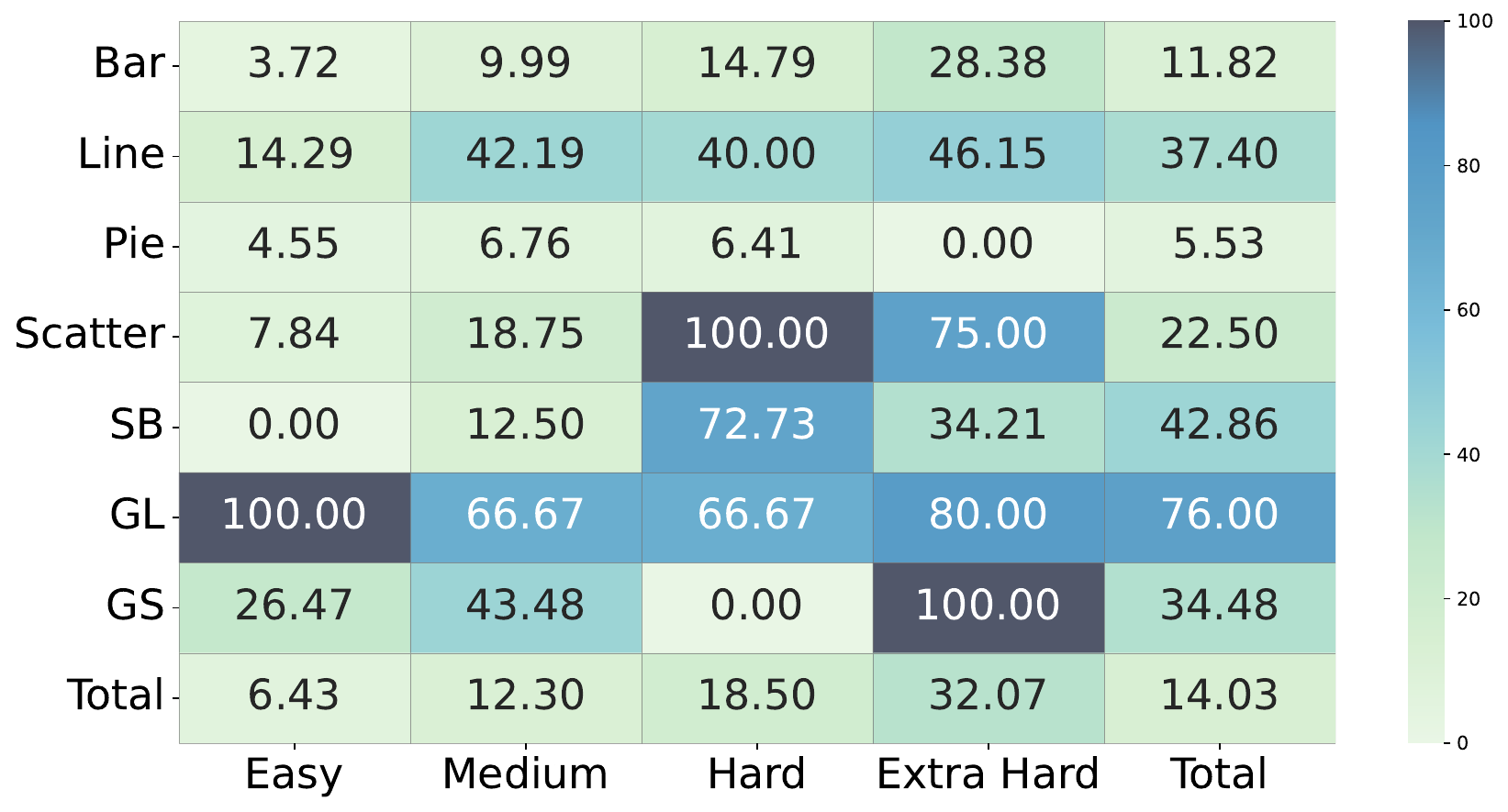}
        \subcaption{Error distribution of \system.}
    \end{minipage}

    \begin{minipage}[b]{\linewidth}
        \centering
        \includegraphics[width=\linewidth]{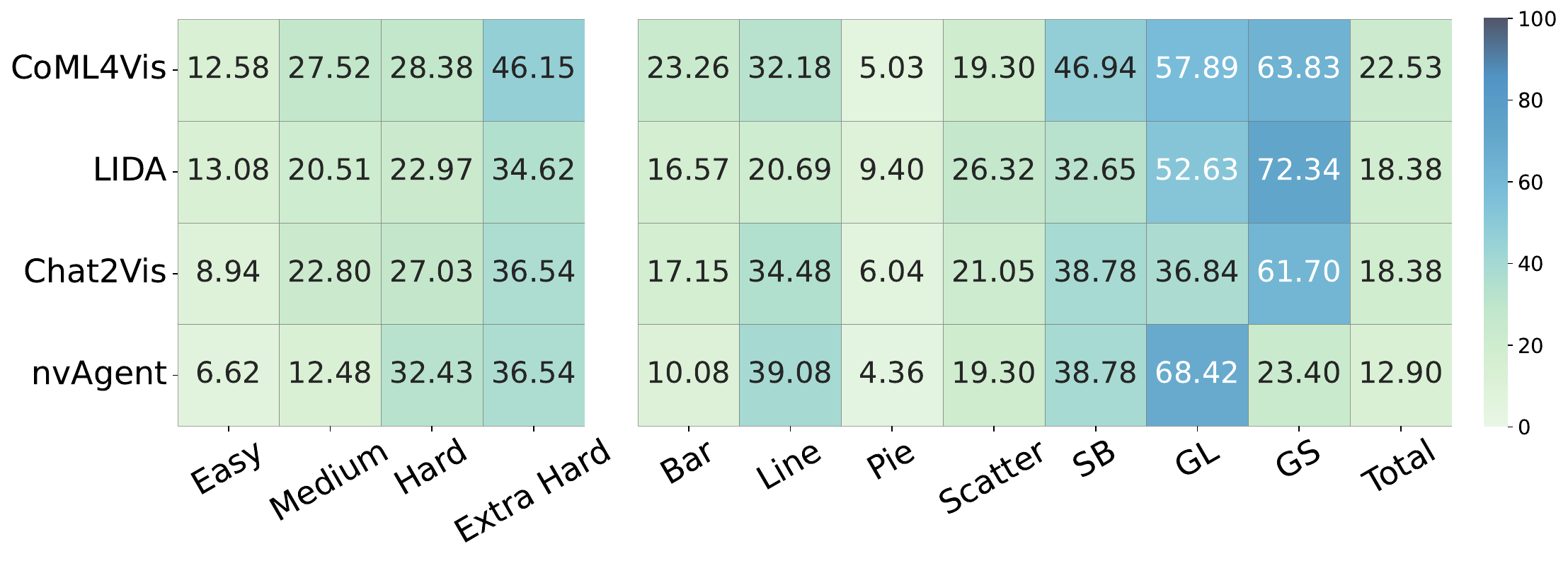}
        \subcaption{Errors of different methods in \textit{single-table} dataset. }
    \end{minipage}

    \caption{Error distributions across hardness levels and chart types. SB, GL, and GS refer to Stacked Bar, Grouping Line, and Grouping Scatter, respectively.}
    \label{fig:Combined_ErrorAnalysis}
\end{figure}

\begin{figure*}[!t]
\centering
\includegraphics[width=\linewidth]{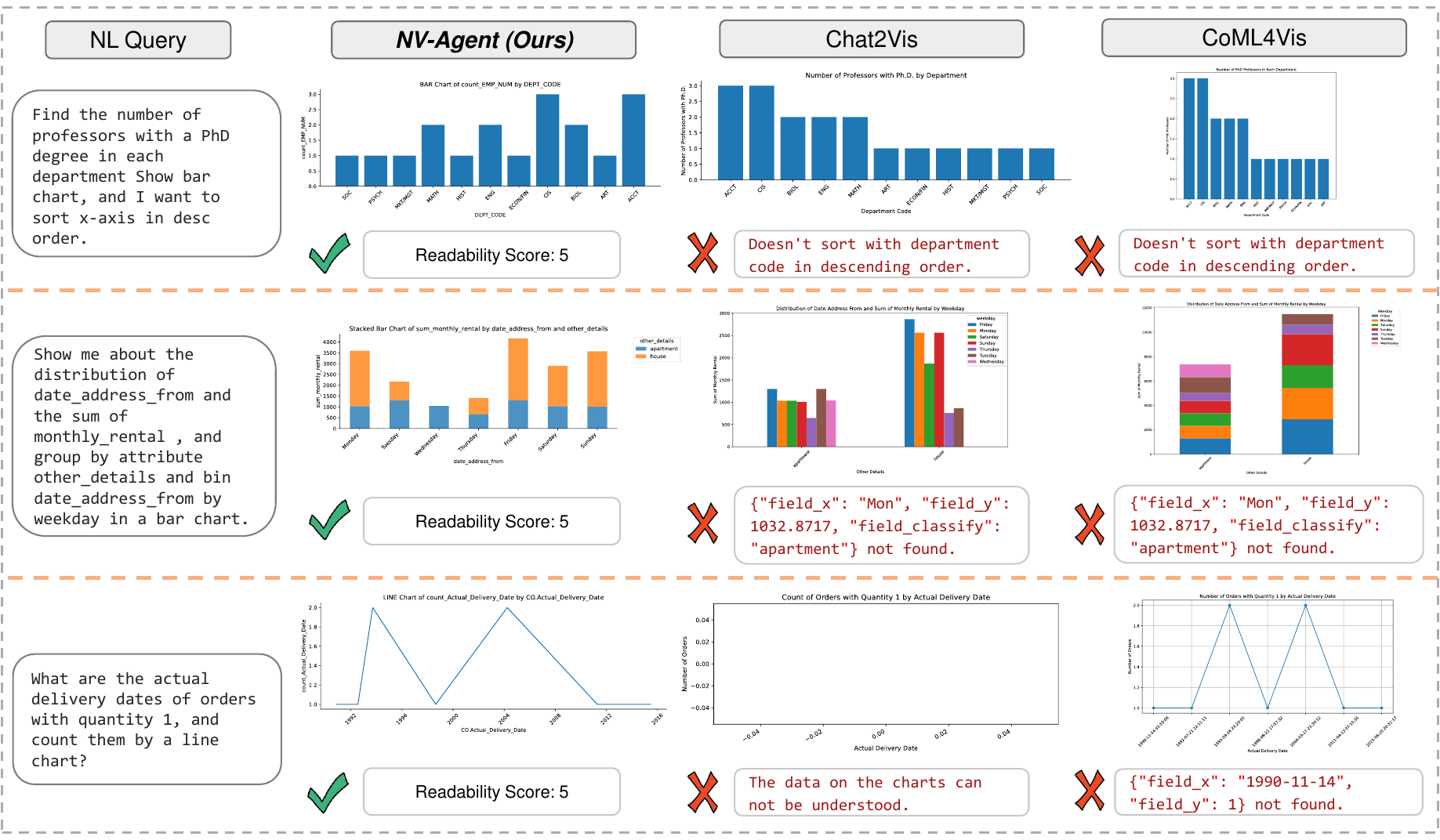}
    \caption{Case study of visualization performed by \system and other baselines. The first two cases are from \textit{single-table} dataset and the third from \textit{multi-table} dataset. \system performed well in most complex cases (\emph{e.g.}, stacked bar charts), while other baselines failed.}
    \label{fig:case_study}
    \vspace{-1em}
\end{figure*}

\paragraph{Error Analysis.} 
As shown in Figure~\ref{fig:Combined_ErrorAnalysis}, \system's performance varies significantly across chart type and difficulty level, particularly with rare queries in temporal data, such as line charts. Our error analysis reveals that failures stem from insufficient handling of temporal information and an imperfect translate function for time-series binning operations. These challenges related to chart complexity and task difficulty underscore the need for better tabular data understanding in LLMs. Our future work can be focused on improving the reasoning abilities of LLMs in temporal information in tabular data. 

\section{Related Work}
\label{related}

\paragraph{\nlvis.}
\nlvis research has evolved from rule-based systems~\citep{nl4dv,orko,flowsense,datatone,deepeye} and neural network-based approaches~\citep{advisor,seq2vis,rgvisnet}, to most recently to generated model enhanced systems~\citep{hong2024data}. 
Current LLM-based approaches can be broadly categorized into two groups: (1) those utilizing prompt engineering techniques, such as Chat2Vis~\citep{chat2vis}, Prompt4Vis~\citep{prompt4vis}, Mirror~\citep{mirror}, LIDA~\citep{lida}, and Data Formulator~\citep{formulator}, and (2) those involving fine-tuning of models specifically for \nlvis tasks, like TableGPT~\citep{tablegpt,tablegpt2}, ChartLlama~\citep{chartllama} and DataVis-T5~\citep{datavist5}. This evolution marks significant progress in making data visualization more accessible and intuitive.

\paragraph{LLM for Tabular Data.}
LLM-based approaches push the performance of tabular data processing to a new boundary~\citep{sqlsurvey}. The emergent in-context learning capability~\citep{icl} and chain-of-thought reasoning~\citep{cot} have significantly enhanced LLMs' ability to handle complex tabular tasks by mimicking examples and encouraging step-by-step thinking~\citep{rethinking,cotautomatic,wu2024tablebench}. These advancements have been particularly impactful in several key tasks such as TableQA~\citep{qiu2024tqa,xu2024llasalargelanguagestructured}, Text2SQL~\citep{rbsql,dinsql}, NL2Formula~\citep{nl2formula} and \nlvis~\citep{cotvis,prompt4vis,chartgpt}. 

\paragraph{Agentic Workflow.}
Agentic workflow leverages multiple LLM-based agents, each assigned different roles to tackle complex problems~\citep{multi}. These systems employ various interaction modes, such as collaboration~\citep{chateval,li2023camel,wu2023autogen} or competition~\citep{zhao2023competeai}, showing remarkable success in database query tasks \citep{macsql,zhu2024autotqa,cen2024sqlfixagent}, software development~\citep{metagpt,mapcoder,agentcoder} and mathematical reasoning~\citep{magicore}. This success stems from the synergy of specialized agents working together to overcome individual limitations and solve complex tasks efficiently.

\section{Conclusion}
\label{conclusion}
In this paper, we have proposed \system, a collaborative agent-based workflow to solve the challenging mult-table \nlvis task and provide a \textit{``turnkey solution''} for users. \system decomposes the process into atomic modules such as database preprocessing, visualization planning, and iterative optimization. 
Experimental results show that \system outperforms state-of-the-art baselines by 7.88\% in single-table and 9.23\% in multi-table scenarios, demonstrating the efficacy of \system.


\newpage

\section*{Limitations}
\label{limitations}
While \system demonstrates significant improvements in \nlvis tasks, we acknowledge several limitations. Our reliance on proprietary APIs may constrain the system's reproducibility and adaptability. Additionally, utilizing large language models as both backbone and evaluator introduces potential biases that could affect output quality and evaluation accuracy. 
Moreover, our error analysis finds insufficient handling of temporal information, which underscores the need for better tabular data understanding capabilities of LLMs. 
Our prompting strategy and evaluation metrics may not fully capture the nuances of complex visualizations or semantic correctness. 
The current framework employs a simple function to translate VQL into Python code, which can be further optimized for better readability.
Future work should address these limitations by exploring open-source alternatives, developing more sophisticated prompting and evaluation techniques, and integrating advanced tools like retrieval augmented generation to enhance the system's capabilities and mitigate biases.

\bibliography{custom}

\begin{thebibliography}{66}
\expandafter\ifx\csname natexlab\endcsname\relax\def\natexlab#1{#1}\fi

\bibitem[{Barrett et~al.(2005)Barrett, Hunter, Miller, Hsu, and Greenfield}]{barrett2005matplotlib}
Paul Barrett, John Hunter, J~Todd Miller, J-C Hsu, and Perry Greenfield. 2005.
\newblock matplotlib--a portable python plotting package.
\newblock In \emph{Astronomical data analysis software and systems XIV}, volume 347, page~91.

\bibitem[{Cen et~al.(2024)Cen, Liu, Li, and Wang}]{cen2024sqlfixagent}
Jipeng Cen, Jiaxin Liu, Zhixu Li, and Jingjing Wang. 2024.
\newblock Sqlfixagent: Towards semantic-accurate sql generation via multi-agent collaboration.
\newblock \emph{arXiv preprint arXiv:2406.13408}.

\bibitem[{Chan et~al.(2023)Chan, Chen, Su, Yu, Xue, Zhang, Fu, and Liu}]{chateval}
Chi-Min Chan, Weize Chen, Yusheng Su, Jianxuan Yu, Wei Xue, Shanghang Zhang, Jie Fu, and Zhiyuan Liu. 2023.
\newblock Chateval: Towards better llm-based evaluators through multi-agent debate.
\newblock \emph{arXiv preprint arXiv:2308.07201}.

\bibitem[{Chen et~al.(2024{\natexlab{a}})Chen, Chen, Zhang, Liu, Wang, Zhou, Zhang, Wan, Zhou, and Sun}]{chen2024mllm}
Dongping Chen, Ruoxi Chen, Shilin Zhang, Yinuo Liu, Yaochen Wang, Huichi Zhou, Qihui Zhang, Yao Wan, Pan Zhou, and Lichao Sun. 2024{\natexlab{a}}.
\newblock Mllm-as-a-judge: Assessing multimodal llm-as-a-judge with vision-language benchmark.
\newblock \emph{arXiv preprint arXiv:2402.04788}.

\bibitem[{Chen et~al.(2024{\natexlab{b}})Chen, Prasad, Saha, Stengel-Eskin, and Bansal}]{magicore}
Justin Chih-Yao Chen, Archiki Prasad, Swarnadeep Saha, Elias Stengel-Eskin, and Mohit Bansal. 2024{\natexlab{b}}.
\newblock \href {http://arxiv.org/abs/2409.12147} {Magicore: Multi-agent, iterative, coarse-to-fine refinement for reasoning}.

\bibitem[{Chen et~al.(2024{\natexlab{c}})Chen, Zhang, Xu, Ren, and Yang}]{viseval}
Nan Chen, Yuge Zhang, Jiahang Xu, Kan Ren, and Yuqing Yang. 2024{\natexlab{c}}.
\newblock Viseval: A benchmark for data visualization in the era of large language models.
\newblock \emph{IEEE Transactions on Visualization and Computer Graphics}.

\bibitem[{Dibia(2023)}]{lida}
Victor Dibia. 2023.
\newblock Lida: A tool for automatic generation of grammar-agnostic visualizations and infographics using large language models.
\newblock \emph{arXiv preprint arXiv:2303.02927}.

\bibitem[{Dong et~al.(2022)Dong, Li, Dai, Zheng, Ma, Li, Xia, Xu, Wu, Liu et~al.}]{icl}
Qingxiu Dong, Lei Li, Damai Dai, Ce~Zheng, Jingyuan Ma, Rui Li, Heming Xia, Jingjing Xu, Zhiyong Wu, Tianyu Liu, et~al. 2022.
\newblock A survey on in-context learning.
\newblock \emph{arXiv preprint arXiv:2301.00234}.

\bibitem[{Elo and Sloan(1978)}]{elo1978rating}
Arpad~E Elo and Sam Sloan. 1978.
\newblock The rating of chessplayers: Past and present.
\newblock \emph{(No Title)}.

\bibitem[{Fiorina()}]{fiorinagoal}
Carly Fiorina.
\newblock The goal is to turn data into information, and information into insight.

\bibitem[{Gao et~al.(2015)Gao, Dontcheva, Adar, Liu, and Karahalios}]{datatone}
Tong Gao, Mira Dontcheva, Eytan Adar, Zhicheng Liu, and Karrie~G Karahalios. 2015.
\newblock Datatone: Managing ambiguity in natural language interfaces for data visualization.
\newblock In \emph{Proceedings of the 28th annual acm symposium on user interface software \& technology}, pages 489--500.

\bibitem[{Han et~al.(2023)Han, Zhang, Chen, Yang, Wang, Yu, Fu, and Zhang}]{chartllama}
Yucheng Han, Chi Zhang, Xin Chen, Xu~Yang, Zhibin Wang, Gang Yu, Bin Fu, and Hanwang Zhang. 2023.
\newblock Chartllama: A multimodal llm for chart understanding and generation.
\newblock \emph{arXiv preprint arXiv:2311.16483}.

\bibitem[{Hong et~al.(2024)Hong, Lin, Liu, Wu, Li, Chen, Zhang, Wang, Zhang, Zhuge et~al.}]{hong2024data}
Sirui Hong, Yizhang Lin, Bangbang Liu, Binhao Wu, Danyang Li, Jiaqi Chen, Jiayi Zhang, Jinlin Wang, Lingyao Zhang, Mingchen Zhuge, et~al. 2024.
\newblock Data interpreter: An llm agent for data science.
\newblock \emph{arXiv preprint arXiv:2402.18679}.

\bibitem[{Hong et~al.(2023)Hong, Zheng, Chen, Cheng, Wang, Zhang, Wang, Yau, Lin, Zhou et~al.}]{metagpt}
Sirui Hong, Xiawu Zheng, Jonathan Chen, Yuheng Cheng, Jinlin Wang, Ceyao Zhang, Zili Wang, Steven Ka~Shing Yau, Zijuan Lin, Liyang Zhou, et~al. 2023.
\newblock Metagpt: Meta programming for multi-agent collaborative framework.
\newblock \emph{arXiv preprint arXiv:2308.00352}.

\bibitem[{Huang et~al.(2024)Huang, Zhang, Luck, Bu, Qing, and Cui}]{agentcoder}
Dong Huang, Jie~M. Zhang, Michael Luck, Qingwen Bu, Yuhao Qing, and Heming Cui. 2024.
\newblock \href {http://arxiv.org/abs/2312.13010} {Agentcoder: Multi-agent-based code generation with iterative testing and optimisation}.

\bibitem[{Islam et~al.(2024)Islam, Ali, and Parvez}]{mapcoder}
Md.~Ashraful Islam, Mohammed~Eunus Ali, and Md~Rizwan Parvez. 2024.
\newblock \href {http://arxiv.org/abs/2405.11403} {Mapcoder: Multi-agent code generation for competitive problem solving}.

\bibitem[{Khan(2024)}]{khan2024data}
Arshad Khan. 2024.
\newblock Data visualization.
\newblock In \emph{Visual Analytics for Dashboards: A Step-by-Step Guide to Principles and Practical Techniques}, pages 67--73. Springer.

\bibitem[{Li et~al.(2023)Li, Hammoud, Itani, Khizbullin, and Ghanem}]{li2023camel}
Guohao Li, Hasan Hammoud, Hani Itani, Dmitrii Khizbullin, and Bernard Ghanem. 2023.
\newblock Camel: Communicative agents for" mind" exploration of large language model society.
\newblock \emph{Advances in Neural Information Processing Systems}, 36:51991--52008.

\bibitem[{Li et~al.(2024{\natexlab{a}})Li, Wang, Aodeng, Zheng, Zhang, Ou, Wang, and Liu}]{li2024visualizationgenerationlargelanguage}
Guozheng Li, Xinyu Wang, Gerile Aodeng, Shunyuan Zheng, Yu~Zhang, Chuangxin Ou, Song Wang, and Chi~Harold Liu. 2024{\natexlab{a}}.
\newblock \href {http://arxiv.org/abs/2401.11255} {Visualization generation with large language models: An evaluation}.

\bibitem[{Li et~al.(2024{\natexlab{b}})Li, Chen, Song, Song, and Zhang}]{prompt4vis}
Shuaimin Li, Xuanang Chen, Yuanfeng Song, Yunze Song, and Chen Zhang. 2024{\natexlab{b}}.
\newblock Prompt4vis: Prompting large language models with example mining and schema filtering for tabular data visualization.
\newblock \emph{arXiv preprint arXiv:2402.07909}.

\bibitem[{Liu et~al.(2024)Liu, Shen, Li, Ma, Jiang, Luo, Zhang, Fan, Li, and Tang}]{sqlsurvey}
Xinyu Liu, Shuyu Shen, Boyan Li, Peixian Ma, Runzhi Jiang, Yuyu Luo, Yuxin Zhang, Ju~Fan, Guoliang Li, and Nan Tang. 2024.
\newblock A survey of nl2sql with large language models: Where are we, and where are we going?
\newblock \emph{arXiv preprint arXiv:2408.05109}.

\bibitem[{Lu et~al.(2024)Lu, Zhang, Fan, Fu, Chen, and Du}]{lu2024large}
Weizheng Lu, Jing Zhang, Ju~Fan, Zihao Fu, Yueguo Chen, and Xiaoyong Du. 2024.
\newblock Large language model for table processing: A survey.
\newblock \emph{arXiv preprint arXiv:2402.05121}.

\bibitem[{Luo et~al.(2018)Luo, Qin, Tang, and Li}]{deepeye}
Yuyu Luo, Xuedi Qin, Nan Tang, and Guoliang Li. 2018.
\newblock Deepeye: Towards automatic data visualization.
\newblock In \emph{2018 IEEE 34th international conference on data engineering (ICDE)}, pages 101--112. IEEE.

\bibitem[{Luo et~al.(2021{\natexlab{a}})Luo, Tang, and Li}]{nvbench}
Yuyu Luo, Jiawei Tang, and Guoliang Li. 2021{\natexlab{a}}.
\newblock nvbench: A large-scale synthesized dataset for cross-domain natural language to visualization task.
\newblock \emph{arXiv preprint arXiv:2112.12926}.

\bibitem[{Luo et~al.(2021{\natexlab{b}})Luo, Tang, Li, Chai, Li, and Qin}]{nvBench_SIGMOD21}
Yuyu Luo, Nan Tang, Guoliang Li, Chengliang Chai, Wenbo Li, and Xuedi Qin. 2021{\natexlab{b}}.
\newblock Synthesizing natural language to visualization (nl2vis) benchmarks from nl2sql benchmarks.
\newblock In \emph{Proceedings of the 2021 International Conference on Management of Data, {SIGMOD} Conference 2021, June 20–25, 2021, Virtual Event, China}. {ACM}.

\bibitem[{Luo et~al.(2021{\natexlab{c}})Luo, Tang, Li, Chai, Li, and Qin}]{seq2vis}
Yuyu Luo, Nan Tang, Guoliang Li, Chengliang Chai, Wenbo Li, and Xuedi Qin. 2021{\natexlab{c}}.
\newblock Synthesizing natural language to visualization (nl2vis) benchmarks from nl2sql benchmarks.
\newblock In \emph{Proceedings of the 2021 International Conference on Management of Data}, pages 1235--1247.

\bibitem[{Maamari et~al.(2024)Maamari, Abubaker, Jaroslawicz, and Mhedhbi}]{death}
Karime Maamari, Fadhil Abubaker, Daniel Jaroslawicz, and Amine Mhedhbi. 2024.
\newblock The death of schema linking? text-to-sql in the age of well-reasoned language models.
\newblock \emph{arXiv preprint arXiv:2408.07702}.

\bibitem[{Maddigan and Susnjak(2023)}]{chat2vis}
Paula Maddigan and Teo Susnjak. 2023.
\newblock Chat2vis: generating data visualizations via natural language using chatgpt, codex and gpt-3 large language models.
\newblock \emph{Ieee Access}, 11:45181--45193.

\bibitem[{Markel et~al.(2002)Markel, Brooker, Hendricks, Johnson, Kelly, Kramer, O’Keefe, Sprik, and Wipke}]{advisor}
Tony Markel, Aaron Brooker, Terry Hendricks, Valerie Johnson, Kenneth Kelly, Bill Kramer, Michael O’Keefe, Sam Sprik, and Keith Wipke. 2002.
\newblock Advisor: a systems analysis tool for advanced vehicle modeling.
\newblock \emph{Journal of power sources}, 110(2):255--266.

\bibitem[{Min et~al.(2022)Min, Lyu, Holtzman, Artetxe, Lewis, Hajishirzi, and Zettlemoyer}]{rethinking}
Sewon Min, Xinxi Lyu, Ari Holtzman, Mikel Artetxe, Mike Lewis, Hannaneh Hajishirzi, and Luke Zettlemoyer. 2022.
\newblock Rethinking the role of demonstrations: What makes in-context learning work?
\newblock \emph{arXiv preprint arXiv:2202.12837}.

\bibitem[{Narechania et~al.(2020)Narechania, Srinivasan, and Stasko}]{nl4dv}
Arpit Narechania, Arjun Srinivasan, and John Stasko. 2020.
\newblock Nl4dv: A toolkit for generating analytic specifications for data visualization from natural language queries.
\newblock \emph{IEEE Transactions on Visualization and Computer Graphics}, 27(2):369--379.

\bibitem[{OpenAI(2022)}]{chatgpt3.5}
OpenAI. 2022.
\newblock Chatgpt (gpt-3.5).
\newblock \url{https://openai.com/index/chatgpt/}.

\bibitem[{OpenAI(2024{\natexlab{a}})}]{openai2024gpt4omini}
OpenAI. 2024{\natexlab{a}}.
\newblock Gpt-4o mini: Advancing cost-efficient intelligence.
\newblock \url{https://openai.com/index/gpt-4o-mini-advancing-cost-efficient-intelligence/}.
\newblock Accessed: 2024-09-04.

\bibitem[{OpenAI(2024{\natexlab{b}})}]{openai_gpt4o_2024}
OpenAI. 2024{\natexlab{b}}.
\newblock \href {https://openai.com/index/hello-gpt-4o/} {Hello gpt-4o}.
\newblock Accessed: 2024-06-06.

\bibitem[{Pourreza and Rafiei(2024)}]{dinsql}
Mohammadreza Pourreza and Davood Rafiei. 2024.
\newblock Din-sql: Decomposed in-context learning of text-to-sql with self-correction.
\newblock \emph{Advances in Neural Information Processing Systems}, 36.

\bibitem[{Qiu et~al.(2024)Qiu, Peng, He, Yuan, and Wang}]{qiu2024tqa}
Zipeng Qiu, You Peng, Guangxin He, Binhang Yuan, and Chen Wang. 2024.
\newblock Tqa-bench: Evaluating llms for multi-table question answering with scalable context and symbolic extension.
\newblock \emph{arXiv preprint arXiv:2411.19504}.

\bibitem[{Sah et~al.(2024)Sah, Mitra, Narechania, Endert, Stasko, and Dou}]{sah2024generating}
Subham Sah, Rishab Mitra, Arpit Narechania, Alex Endert, John Stasko, and Wenwen Dou. 2024.
\newblock Generating analytic specifications for data visualization from natural language queries using large language models.
\newblock \emph{arXiv preprint arXiv:2408.13391}.

\bibitem[{Song et~al.(2022)Song, Zhao, Wong, and Jiang}]{rgvisnet}
Yuanfeng Song, Xuefang Zhao, Raymond Chi-Wing Wong, and Di~Jiang. 2022.
\newblock Rgvisnet: A hybrid retrieval-generation neural framework towards automatic data visualization generation.
\newblock In \emph{Proceedings of the 28th ACM SIGKDD Conference on Knowledge Discovery and Data Mining}, pages 1646--1655.

\bibitem[{Srinivasan and Stasko(2017)}]{orko}
Arjun Srinivasan and John Stasko. 2017.
\newblock Orko: Facilitating multimodal interaction for visual exploration and analysis of networks.
\newblock \emph{IEEE transactions on visualization and computer graphics}, 24(1):511--521.

\bibitem[{Su et~al.(2024)Su, Wang, Ye, Zhou, Zhang, Zhu, Wang, Xu, Chen, Li, Lan, Tian, Yuan, Zhao, Zhou, Shou, Zha, Long, Li, Wu, Zhang, Huang, Yang, Zhang, Ye, Zhu, Hu, Gu, Sun, Li, Yang, and Xiao}]{tablegpt2}
Aofeng Su, Aowen Wang, Chao Ye, Chen Zhou, Ga~Zhang, Guangcheng Zhu, Haobo Wang, Haokai Xu, Hao Chen, Haoze Li, Haoxuan Lan, Jiaming Tian, Jing Yuan, Junbo Zhao, Junlin Zhou, Kaizhe Shou, Liangyu Zha, Lin Long, Liyao Li, Pengzuo Wu, Qi~Zhang, Qingyi Huang, Saisai Yang, Tao Zhang, Wentao Ye, Wufang Zhu, Xiaomeng Hu, Xijun Gu, Xinjie Sun, Xiang Li, Yuhang Yang, and Zhiqing Xiao. 2024.
\newblock \href {http://arxiv.org/abs/2411.02059} {Tablegpt2: A large multimodal model with tabular data integration}.

\bibitem[{Sui et~al.(2024)Sui, Zhou, Zhou, Han, and Zhang}]{augment}
Yuan Sui, Mengyu Zhou, Mingjie Zhou, Shi Han, and Dongmei Zhang. 2024.
\newblock Table meets llm: Can large language models understand structured table data? a benchmark and empirical study.
\newblock In \emph{Proceedings of the 17th ACM International Conference on Web Search and Data Mining}, pages 645--654.

\bibitem[{Talebirad and Nadiri(2023)}]{multi}
Yashar Talebirad and Amirhossein Nadiri. 2023.
\newblock Multi-agent collaboration: Harnessing the power of intelligent llm agents.
\newblock \emph{arXiv preprint arXiv:2306.03314}.

\bibitem[{Tian et~al.(2024)Tian, Cui, Deng, Yi, Yang, Zhang, and Wu}]{chartgpt}
Yuan Tian, Weiwei Cui, Dazhen Deng, Xinjing Yi, Yurun Yang, Haidong Zhang, and Yingcai Wu. 2024.
\newblock Chartgpt: Leveraging llms to generate charts from abstract natural language.
\newblock \emph{IEEE Transactions on Visualization and Computer Graphics}.

\bibitem[{Vartak et~al.(2017)Vartak, Huang, Siddiqui, Madden, and Parameswaran}]{towardsvisualization}
Manasi Vartak, Silu Huang, Tarique Siddiqui, Samuel Madden, and Aditya Parameswaran. 2017.
\newblock Towards visualization recommendation systems.
\newblock \emph{Acm Sigmod Record}, 45(4):34--39.

\bibitem[{Wan et~al.(2024)Wan, Song, Li, Zhang, and Wong}]{datavist5}
Zhuoyue Wan, Yuanfeng Song, Shuaimin Li, Chen~Jason Zhang, and Raymond Chi-Wing Wong. 2024.
\newblock Datavist5: A pre-trained language model for jointly understanding text and data visualization.
\newblock \emph{arXiv preprint arXiv:2408.07401}.

\bibitem[{Wang et~al.(2024{\natexlab{a}})Wang, Ren, Yang, Liang, Bai, Chai, Yan, Zhang, Yin, Sun, and Li}]{macsql}
Bing Wang, Changyu Ren, Jian Yang, Xinnian Liang, Jiaqi Bai, Linzheng Chai, Zhao Yan, Qian-Wen Zhang, Di~Yin, Xing Sun, and Zhoujun Li. 2024{\natexlab{a}}.
\newblock \href {http://arxiv.org/abs/2312.11242} {Mac-sql: A multi-agent collaborative framework for text-to-sql}.

\bibitem[{Wang et~al.(2024{\natexlab{b}})Wang, Thompson, and Lee}]{formulator}
Chenglong Wang, John Thompson, and Bongshin Lee. 2024{\natexlab{b}}.
\newblock \href {https://doi.org/10.1109/TVCG.2023.3326585} {Data formulator: Ai-powered concept-driven visualization authoring}.
\newblock \emph{IEEE Transactions on Visualization and Computer Graphics}, 30(1):1128--1138.

\bibitem[{Waskom(2021)}]{seaborn}
Michael~L Waskom. 2021.
\newblock Seaborn: statistical data visualization.
\newblock \emph{Journal of Open Source Software}, 6(60):3021.

\bibitem[{Wei et~al.(2022{\natexlab{a}})Wei, Wang, Schuurmans, Bosma, Xia, Chi, Le, Zhou et~al.}]{wei2022chain}
Jason Wei, Xuezhi Wang, Dale Schuurmans, Maarten Bosma, Fei Xia, Ed~Chi, Quoc~V Le, Denny Zhou, et~al. 2022{\natexlab{a}}.
\newblock Chain-of-thought prompting elicits reasoning in large language models.
\newblock \emph{Advances in Neural Information Processing Systems}, 35:24824--24837.

\bibitem[{Wei et~al.(2022{\natexlab{b}})Wei, Wang, Schuurmans, Bosma, Xia, Chi, Le, Zhou et~al.}]{cot}
Jason Wei, Xuezhi Wang, Dale Schuurmans, Maarten Bosma, Fei Xia, Ed~Chi, Quoc~V Le, Denny Zhou, et~al. 2022{\natexlab{b}}.
\newblock Chain-of-thought prompting elicits reasoning in large language models.
\newblock \emph{Advances in neural information processing systems}, 35:24824--24837.

\bibitem[{Wu et~al.(2023)Wu, Bansal, Zhang, Wu, Zhang, Zhu, Li, Jiang, Zhang, and Wang}]{wu2023autogen}
Qingyun Wu, Gagan Bansal, Jieyu Zhang, Yiran Wu, Shaokun Zhang, Erkang Zhu, Beibin Li, Li~Jiang, Xiaoyun Zhang, and Chi Wang. 2023.
\newblock Autogen: Enabling next-gen llm applications via multi-agent conversation framework.
\newblock \emph{arXiv preprint arXiv:2308.08155}.

\bibitem[{Wu et~al.(2024{\natexlab{a}})Wu, Yang, Chai, Zhang, Liu, Du, Liang, Shu, Cheng, Sun et~al.}]{wu2024tablebench}
Xianjie Wu, Jian Yang, Linzheng Chai, Ge~Zhang, Jiaheng Liu, Xinrun Du, Di~Liang, Daixin Shu, Xianfu Cheng, Tianzhen Sun, et~al. 2024{\natexlab{a}}.
\newblock Tablebench: A comprehensive and complex benchmark for table question answering.
\newblock \emph{arXiv preprint arXiv:2408.09174}.

\bibitem[{Wu et~al.(2024{\natexlab{b}})Wu, Wan, Zhang, Sui, Wei, Zhao, Xu, and Jin}]{automated}
Yang Wu, Yao Wan, Hongyu Zhang, Yulei Sui, Wucai Wei, Wei Zhao, Guandong Xu, and Hai Jin. 2024{\natexlab{b}}.
\newblock Automated data visualization from natural language via large language models: An exploratory study.
\newblock \emph{Proceedings of the ACM on Management of Data}, 2(3):1--28.

\bibitem[{Wu et~al.(2024{\natexlab{c}})Wu, Li, Zhang, Li, Zhao, Fang, He, Li, Li, and Song}]{rbsql}
Zhenhe Wu, Zhongqiu Li, Jie Zhang, Mengxiang Li, Yu~Zhao, Ruiyu Fang, Zhongjiang He, Xuelong Li, Zhoujun Li, and Shuangyong Song. 2024{\natexlab{c}}.
\newblock Rb-sql: A retrieval-based llm framework for text-to-sql.
\newblock \emph{arXiv preprint arXiv:2407.08273}.

\bibitem[{Xu et~al.(2023)Xu, McAuley, and Wang}]{mirror}
Canwen Xu, Julian McAuley, and Penghan Wang. 2023.
\newblock \href {https://doi.org/10.1145/3543873.3587309} {Mirror: A natural language interface for data querying, summarization, and visualization}.
\newblock In \emph{Companion Proceedings of the ACM Web Conference 2023}, WWW ’23, page 49–52. ACM.

\bibitem[{Xu et~al.(2024)Xu, He, Xiangrong, Chen, Liu, Wang, Zhao, and Liu}]{xu2024llasalargelanguagestructured}
Yao Xu, Shizhu He, Zeng Xiangrong, Jiabei Chen, Guang Liu, Bingning Wang, Jun Zhao, and Kang Liu. 2024.
\newblock \href {http://arxiv.org/abs/2411.14460} {Llasa: Large language and structured data assistant}.

\bibitem[{Yang et~al.(2024)Yang, Yang, Zhao, Li, and Rao}]{cotvis}
Hao Yang, Zhaoyong Yang, Ruyang Zhao, Xiaoran Li, and Gaoqi Rao. 2024.
\newblock The implementation solution for automatic visualization of tabular data in relational databases based on large language models.
\newblock In \emph{2024 International Conference on Asian Language Processing (IALP)}, pages 175--180. IEEE.

\bibitem[{Ye et~al.(2024)Ye, Wang, Huang, Chen, Zhang, Moniz, Gao, Geyer, Huang, Chen et~al.}]{ye2024justice}
Jiayi Ye, Yanbo Wang, Yue Huang, Dongping Chen, Qihui Zhang, Nuno Moniz, Tian Gao, Werner Geyer, Chao Huang, Pin-Yu Chen, et~al. 2024.
\newblock Justice or prejudice? quantifying biases in llm-as-a-judge.
\newblock \emph{arXiv preprint arXiv:2410.02736}.

\bibitem[{Yin et~al.(2024)Yin, Li, Sun, Ibrar, and Teng}]{yin2024data}
Shoulin Yin, Hang Li, Yang Sun, Muhammad Ibrar, and Lin Teng. 2024.
\newblock Data visualization analysis based on explainable artificial intelligence: A survey.
\newblock \emph{IJLAI Transactions on Science and Engineering}, 2(2):13--20.

\bibitem[{Yu and Silva(2019)}]{flowsense}
Bowen Yu and Cl{\'a}udio~T Silva. 2019.
\newblock Flowsense: A natural language interface for visual data exploration within a dataflow system.
\newblock \emph{IEEE transactions on visualization and computer graphics}, 26(1):1--11.

\bibitem[{Zha et~al.(2023)Zha, Zhou, Li, Wang, Huang, Yang, Yuan, Su, Li, Su et~al.}]{tablegpt}
Liangyu Zha, Junlin Zhou, Liyao Li, Rui Wang, Qingyi Huang, Saisai Yang, Jing Yuan, Changbao Su, Xiang Li, Aofeng Su, et~al. 2023.
\newblock Tablegpt: Towards unifying tables, nature language and commands into one gpt.
\newblock \emph{arXiv preprint arXiv:2307.08674}.

\bibitem[{Zhang et~al.(2023)Zhang, Zhang, Ren, Li, and Yang}]{coml}
Lei Zhang, Yuge Zhang, Kan Ren, Dongsheng Li, and Yuqing Yang. 2023.
\newblock Mlcopilot: Unleashing the power of large language models in solving machine learning tasks.
\newblock \emph{arXiv preprint arXiv:2304.14979}.

\bibitem[{Zhang et~al.(2022)Zhang, Zhang, Li, and Smola}]{cotautomatic}
Zhuosheng Zhang, Aston Zhang, Mu~Li, and Alex Smola. 2022.
\newblock Automatic chain of thought prompting in large language models.
\newblock \emph{arXiv preprint arXiv:2210.03493}.

\bibitem[{Zhao et~al.(2023)Zhao, Wang, Zhang, Jin, Zhu, Chen, and Xie}]{zhao2023competeai}
Qinlin Zhao, Jindong Wang, Yixuan Zhang, Yiqiao Jin, Kaijie Zhu, Hao Chen, and Xing Xie. 2023.
\newblock Competeai: Understanding the competition behaviors in large language model-based agents.
\newblock \emph{arXiv preprint arXiv:2310.17512}.

\bibitem[{Zhao et~al.(2024)Zhao, Hou, Wu, Gao, Dong, Wan, Zhang, Sui, and Zhang}]{nl2formula}
Wei Zhao, Zhitao Hou, Siyuan Wu, Yan Gao, Haoyu Dong, Yao Wan, Hongyu Zhang, Yulei Sui, and Haidong Zhang. 2024.
\newblock Nl2formula: Generating spreadsheet formulas from natural language queries.
\newblock In \emph{Findings of the Association for Computational Linguistics: EACL 2024}, pages 2377--2388.

\bibitem[{Zhu et~al.(2024)Zhu, Cai, Xu, Li, Sun, Zhou, Su, Tang, and Liu}]{zhu2024autotqa}
Jun-Peng Zhu, Peng Cai, Kai Xu, Li~Li, Yishen Sun, Shuai Zhou, Haihuang Su, Liu Tang, and Qi~Liu. 2024.
\newblock Autotqa: Towards autonomous tabular question answering through multi-agent large language models.
\newblock \emph{Proceedings of the VLDB Endowment}, 17(12):3920--3933.

\end{thebibliography}
\bibliographystyle{acl_natbib}

\clearpage
\appendix

\section{Detailed Experiment Setups}
\label{detailed_experiment_setups}
\paragraph{Baselines.}
\label{detailed_baselines}
This study compares our approach with three state-of-the-art baselines. We also attempted to include Code Interpreter as a baseline; however, API rate limitations prevent the direct generation of visualizations from CSV files.

\begin{itemize}[leftmargin=*, itemsep=0pt] 
    \item \textbf{Chat2Vis} \cite{chat2vis}: It generates data visualizations by leveraging prompt engineering to translate natural language descriptions into visualizations. It uses a language-based table description, which includes column types and sample values, to inform the visualization generation process.\item \textbf{LIDA} \cite{lida}: It structures visualization generation as a four-step process, where each step builds on the previous one to incrementally translate natural language inputs into visualizations. It uses a JSON format to describe column statistics and samples, making it adaptable across various visualization tasks.
    \item \textbf{CoML4Vis} \cite{coml}: 
    It utilizes a few-shot prompt that integrates multiple tables into a single visualization task. It summarizes data table information, including column names and samples, and then applies a few-shot prompt to guide visualization generation.
\end{itemize}

\paragraph{Metrics.}
\label{detailed_metrics}
Our evaluation framework involves five main metrics:
\begin{itemize}[leftmargin=*, itemsep=0pt] 
    \item \textbf{Invalid Rate} represents the percentage of visualizations that fail to render due to issues like incorrect API usage or other code errors.
    \item \textbf{Illegal Rate} indicates the percentage of visualizations that do not meet query requirements, which can include incorrect data transformations, mismatched chart types, or improper visualizations.
    \item \textbf{Readability Score} is the average score (range 1-5) assigned by a vision language model, like GPT-4V, for valid and legal visualizations, assessing their visual clarity and ease of interpretation.
    \item \textbf{Pass Rate} measures the proportion of visualizations in the evaluation set that are both valid (able to render) and legal (meet the query requirements).
    \item \textbf{Quality Score} is set to 0 for invalid or illegal visualizations; otherwise, it is equal to the readability score, providing an overall assessment of visualization quality factoring in both functionality and clarity.
\end{itemize}
To thoroughly evaluate each main metric, we further break them down into the following detailed assessment criteria:
\begin{itemize}[leftmargin=4mm, itemsep=0.05mm] 
    \item \textbf{Code Execution Check} verifies that the Python code generated by the model can be successfully executed.
    \item \textbf{Surface-form Check} ensures that the generated code includes necessary elements to produce a visualization like function calls to display the chart.
    \item \textbf{Chart Type Check} verifies whether the extracted chart type from the visualization matches the ground truth.
    \item \textbf{Data Check} assesses if the data used in the visualization matches the ground truth, taking into consideration potential channel swaps based on specified channels.
    \item \textbf{Order Check} evaluates whether the sorting of visual elements follows the specified query requirements.
    \item \textbf{Layout Check} examines issues like text overflow or element overlap within visualizations.
    \item \textbf{Scale \& Ticks Check} ensures that scales and ticks are appropriately chosen, avoiding unconventional representations.
    \item \textbf{Overall Readability Rating} integrates various readability checks to provide a comprehensive score considering layout, scale, text clarity, and arrangement.
\end{itemize}

The evaluation metrics are averaged across the dataset to provide a comprehensive overview of the model's performance. Together, these metrics ensure that the visualizations are both accurate in execution and effective in conveying the intended data narratives.

\begin{table}[!t]
\centering
\setlength{\belowcaptionskip}{0em} 
\begin{tabular}{lcc}
\toprule[1.5pt]
\textbf{Model} & \textbf{P-corr} & \textbf{P-value} \\
\midrule
GPT-4o-mini & \textbf{0.6503} & 0.000 \\
GPT-4o & 0.5648 & 0.000 \\
\bottomrule[1.5pt]
\end{tabular}
\caption{ The Pearson correlations of GPT-4o-mini and GPT-4o with human judgments on readability scores. }
\label{tab:pearson_corr}
\vspace{-1em}
\end{table}

\begin{table*}[!ht]
\centering

\vspace{-1em}
\begin{tabular}{l|ccc|ccc}
\toprule
\multirow{2}{*}{Method} & \multicolumn{3}{c|}{Single Table} & \multicolumn{3}{c}{Multiple Tables} \\
\cmidrule(l){2-4} \cmidrule(l){5-7}
 & prompt & response & total & prompt & response & total \\
\midrule
LIDA & 1386.23 & 237.90 & 1624.13 & \multicolumn{3}{c}{N/A} \\
Chat2Vis & 414.35 & 451.30 & 865.65 & \multicolumn{3}{c}{N/A} \\
CoML4Vis & 2614.76 & 279.86 & 2894.62 & 3069.62 & 307.67 & 3377.29 \\
\system & 5122.99 & 777.63 & 5900.62 & 5613.96 & 1014.10 & 6628.06 \\
\bottomrule
\end{tabular}
\caption{Token usage comparison for different methods. N/A indicates that LIDA and Chat2Vis cannot handle multiple table scenarios.}
\label{tab:token_usage}
\end{table*}

\begin{table}[ht]
\centering
\scalebox{1}{
\begin{tabular}{l|ccc}
\toprule
Agent & \#Input & \#Output & \#Total \\
\midrule
Processor & 1486.07 & 569.58 & 1755.65\\
Composer & 3268.32 & 221.74 & 3490.07 \\
Validator & 1051.82 & 127.85 & 1179.67  \\
\bottomrule
\end{tabular}}
\caption{Token usage of three agents in \system.} \label{tab:token_agent} 
\vspace{-1em}
\end{table}

\paragraph{Implement Details.}
Our system is implemented in Python 3.9, utilizing GPT-4o \citep{openai_gpt4o_2024}, GPT-4o-mini~\cite{openai2024gpt4omini}, and GPT-3.5-turbo~\cite{chatgpt3.5} as the backbone model for all approaches, with the temperature set to 0 for consistent outputs. GPT-4o-mini serves as the vision language model for readability evaluation. We interact with these models through the Azure OpenAI API. The specific prompt templates for each agent, crucial for guiding their respective roles in the visualization generation process, are detailed in Appendix~\ref{prompt_details}. Token usages of \system and baselines are demonstrated in Table~\ref{tab:token_usage}, and usage for each agent in our \system is shown in Table~\ref{tab:token_agent}. Additionally, our evaluations are conducted in VisEval Benchmark (with MIT license).

\begin{figure*}[!h]
    \centering
    \includegraphics[width=0.98\linewidth]{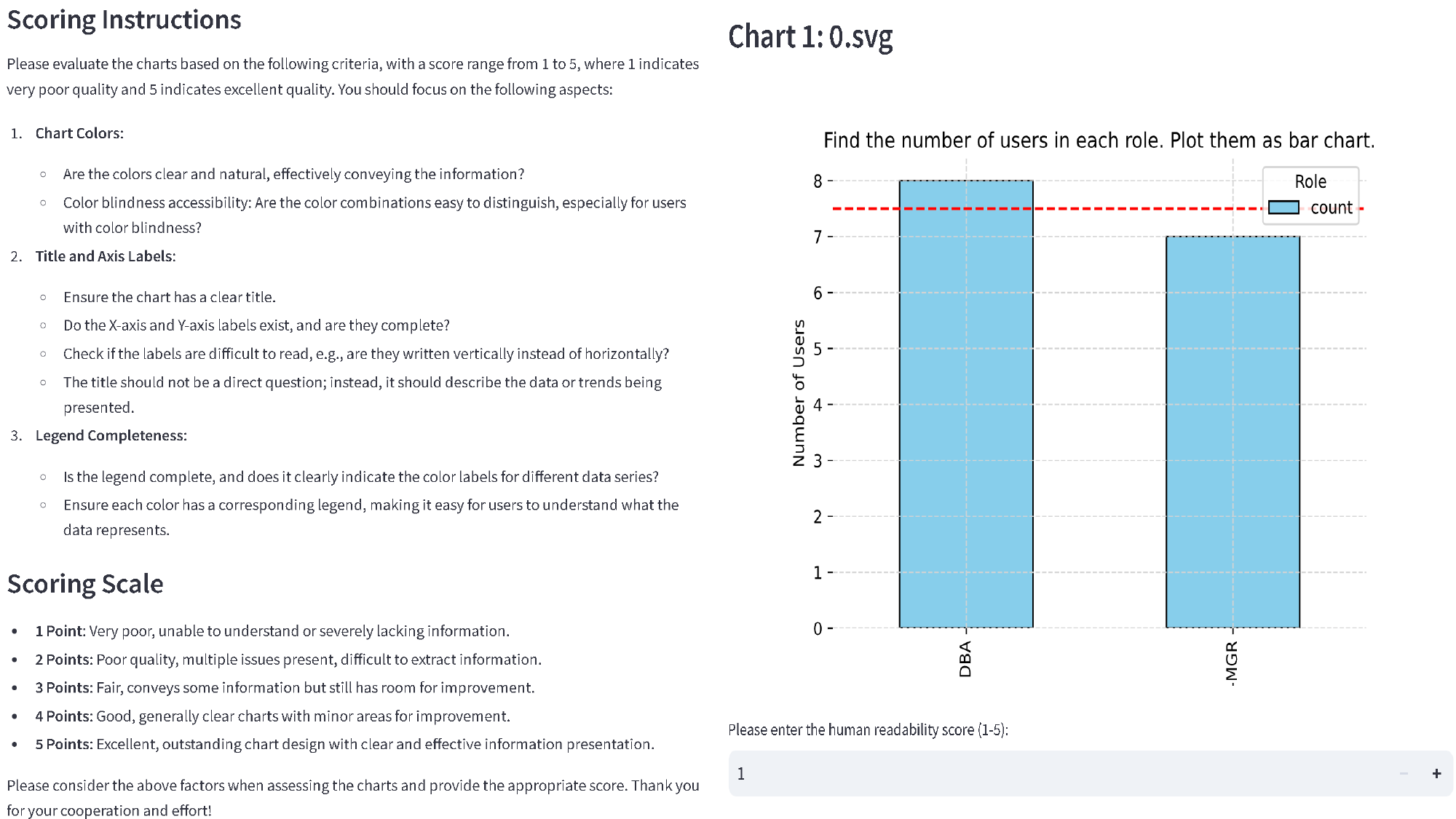}
    \caption{Screenshot of human annotation process in readability score.}
    \label{fig:annotation}
\end{figure*}

\paragraph{Human Annotation.}
\label{human}
The annotation is conducted by 5 authors of this paper independently. As acknowledged, the diversity of annotators plays a crucial role in reducing bias and enhancing the reliability of the benchmark. These annotators have knowledge in the data visualization domain, with different genders, ages, and educational backgrounds. The educational backgrounds of annotators are above undergraduate. To ensure the annotators can proficiently mark the data, we provide them with detailed tutorials, teaching them how to judge the quality of data visualization. We also provide them with detailed criteria and task requirements in each annotation process shown in Figure~\ref{fig:annotation}. Two experiments requiring human annotation are detailed as follows:

\begin{figure}[!ht]
    \centering
    \includegraphics[width=\linewidth]{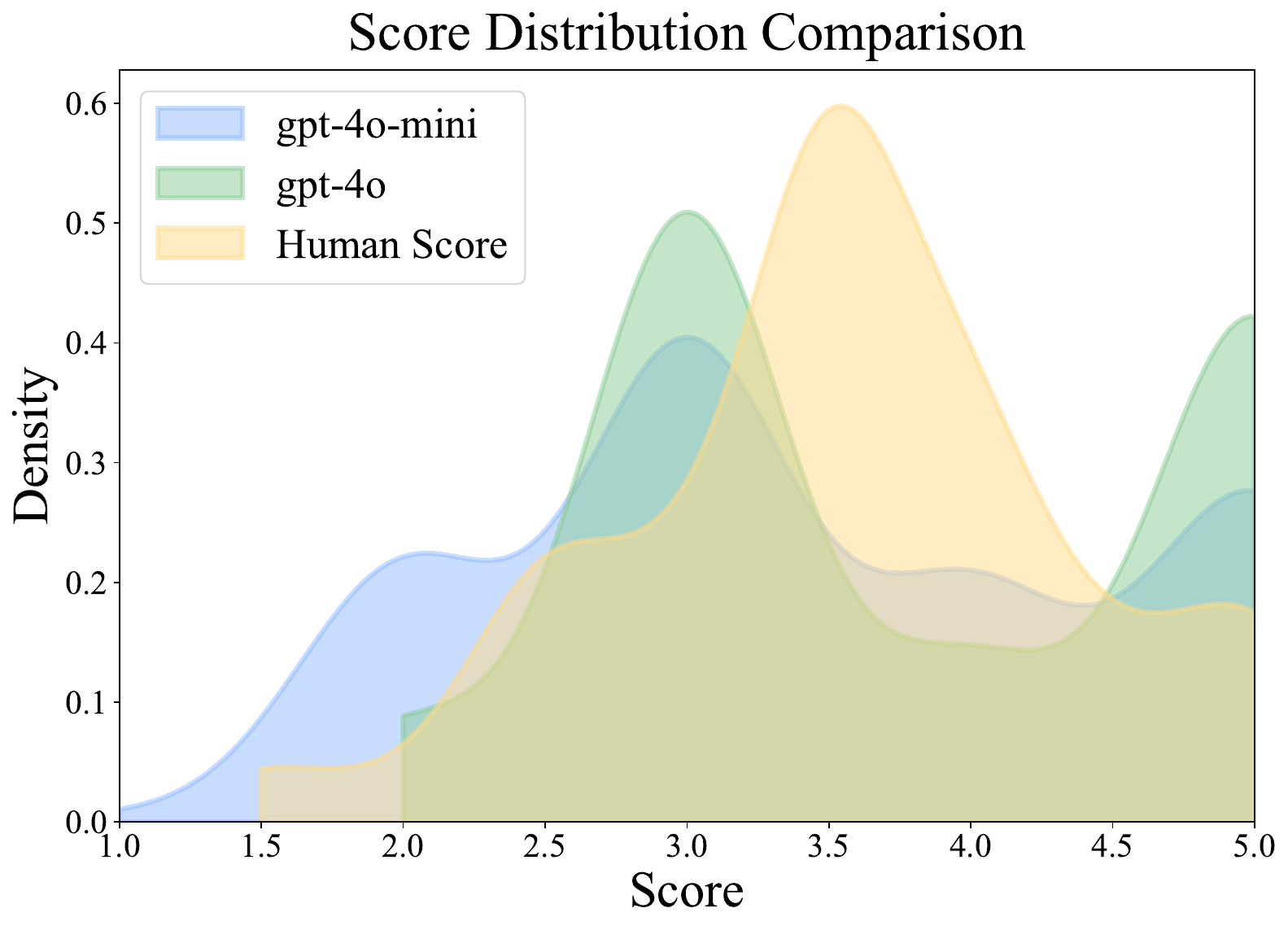}
    \caption{Comparison of score density distribution between GPT-4o, GPT-4o-mini and human average score.}
    \label{fig:score_distribution}
\end{figure}

\begin{table*}[!ht]
\centering
\begin{tabular}{l|ccc}
\toprule
& Invalid Rate & Illegal Rate & Pass Rate \\
\midrule
\system & 4.66\% & 23.97\% & 71.35\% \\
w. CoT for Validator & 5.82\% & 23.39\% & 70.78\% \\
w. original schema for Validator & 4.80\% & 24.22\% & 70.97\% \\
\bottomrule
\end{tabular}
\caption{Additional exploration for Validator (using GPT-3.5-turbo).} 
\vspace{-1em} 
\label{tab:ablation_validator}
\end{table*}

\begin{figure*}[!ht]
    \centering
    \includegraphics[width=0.98\linewidth]{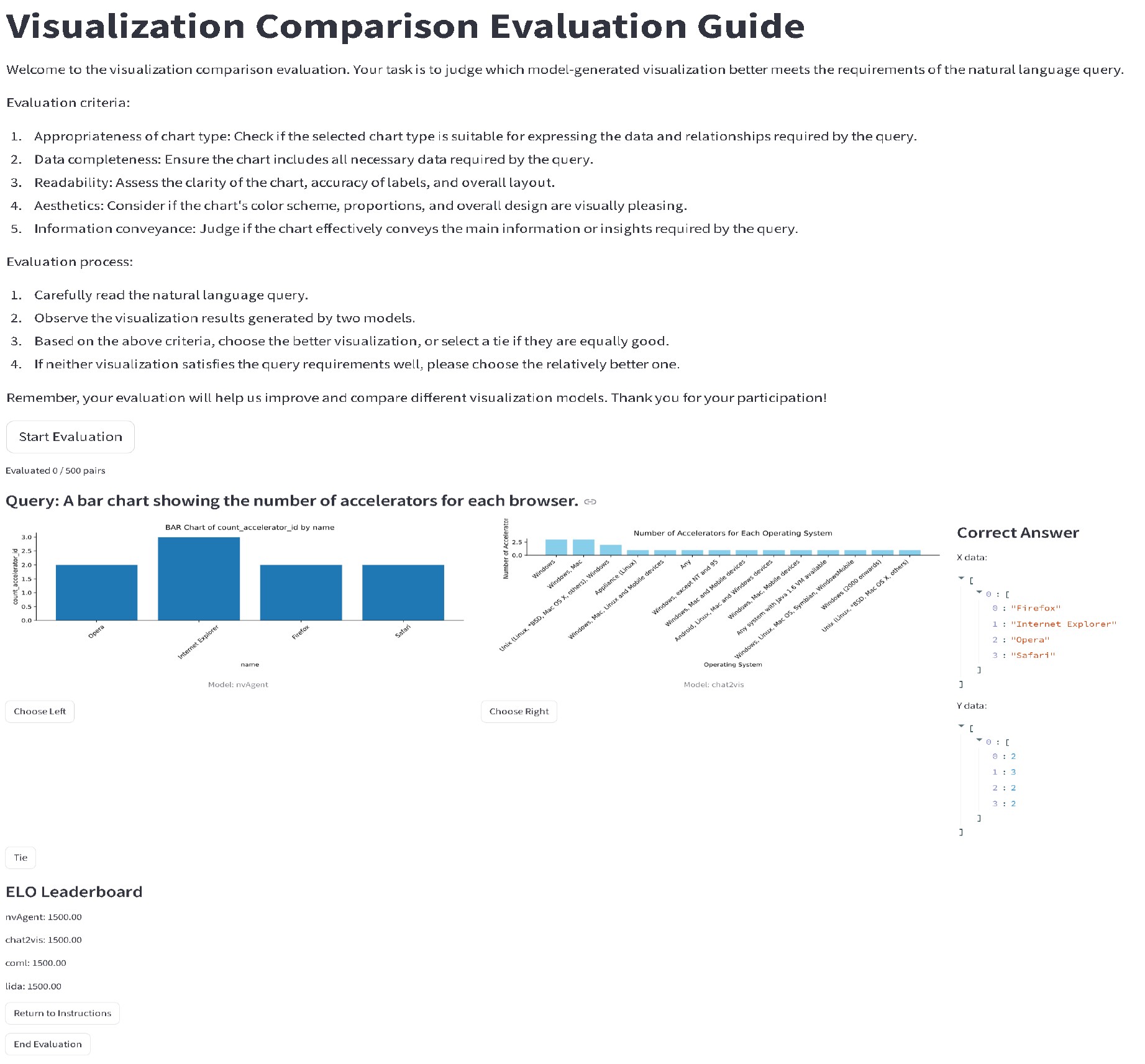}
    \caption{Screenshot of ELO score evaluation framework for Human-as-a-Judge.}
    \label{fig:elo}
\end{figure*}

\begin{itemize}[leftmargin=*, itemsep=0pt]
    \item \textbf{Pearson Correlation of Visual Language Model.} We conduct human annotation frameworks to compare the ability of the visual language model for MLLM-as-a-Judge~\cite{chen2024mllm}, providing the readability score. Our annotation framework is shown in Figure~\ref{fig:annotation}. The final Pearson scores are demonstrated in Table~\ref{tab:pearson_corr}, with its density distribution in Figure~\ref{fig:score_distribution}. The detailed instructions can be found in Figure~\ref{fig:scoring_instructions}.
    \item \textbf{Qualitative comparison to calculate ELO Scores.} We conduct human-judgments evaluations to compare which visualization generated by different models meets the query requirement more precisely. The leaderboard is shown in Table~\ref{tab:elo_rankings}, and Figure~\ref{fig:elo} shows the judgment framework. Each model starts with a base ELO score of 1500. After each pairwise comparison, the scores are updated based on the outcome and the current scores of the models involved. The hyperparameters are set as follows: the $K$-factor is set to 32, which determines the maximum change in rating after a single comparison. We conduct two sets of evaluations: one for single-table queries and another for multiple-table queries, with 1000 bootstrap iterations for each set to ensure statistical robustness. For each model's ELO rating, we report the 95\% confidence intervals computed through bootstrap resampling, providing a measure of rating stability. The evaluation process involves presenting human judges with a query and two visualizations, asking them to select the one that better meets the query requirements. This process is repeated across all model pairs and queries in our test set. The detailed guidance provides to the human evaluators can be found in Figure~\ref{fig:evaluation_instructions}, which outlines the criteria for judging visualization quality and relevance to the given query.

\end{itemize}

\begin{figure}[!ht]
	\centering
    \setlength{\belowcaptionskip}{-1em}
	\includegraphics[width=0.98\linewidth,scale=1.0]
    {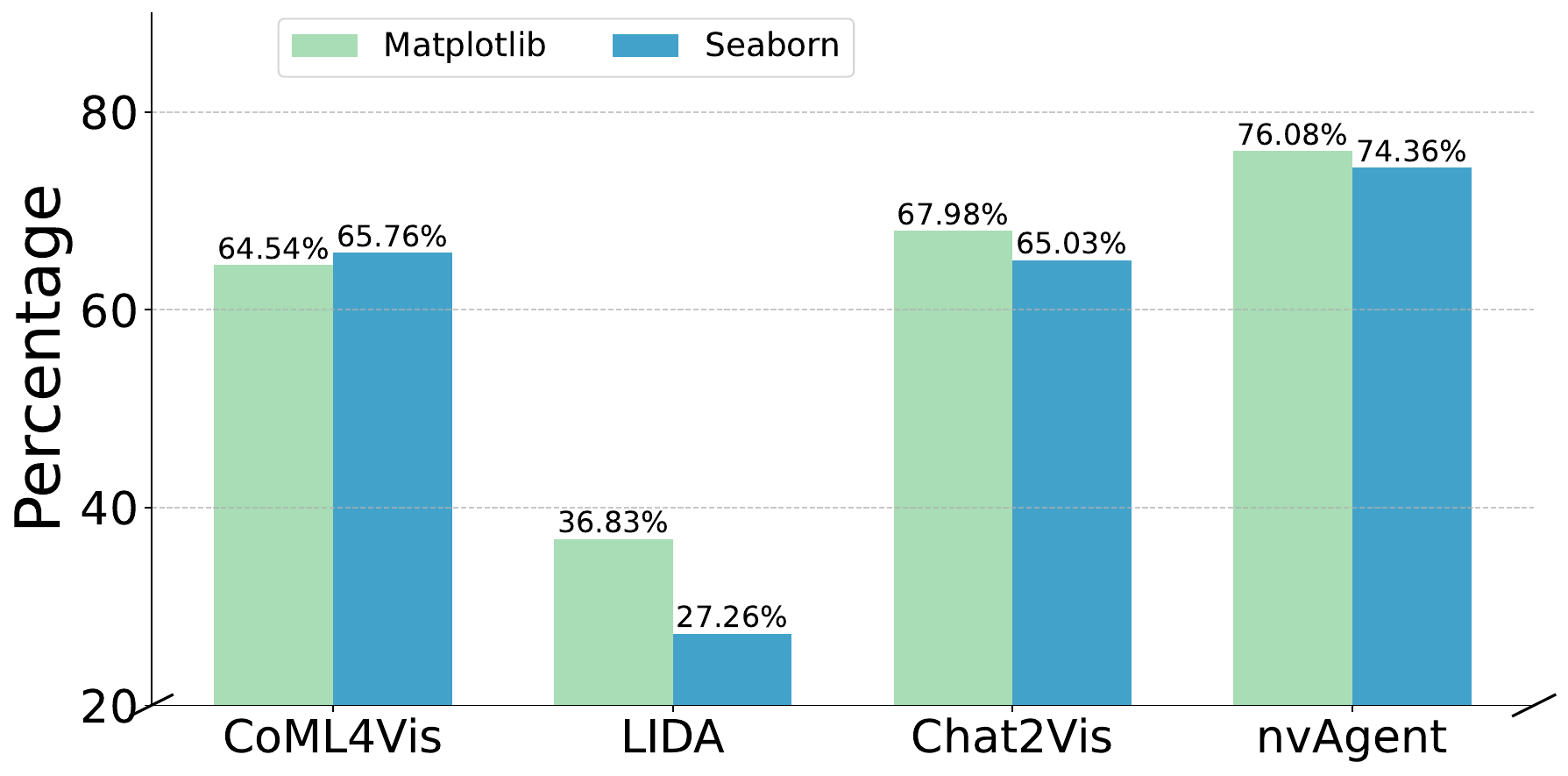}
    \vspace{-1em}
	\caption{Performance of different models using \texttt{Matplotlib} and \texttt{Seaborn} libraries, using GPT-3.5-turbo.
    }
\label{fig: library}
\end{figure}

\begin{figure*}[ht]
\centering
\vspace{1em}
\begin{tcolorbox}[enhanced,attach boxed title to top center={yshift=-3mm,yshifttext=-1mm},boxrule=0.9pt, 
  colback=gray!00,colframe=black!50,colbacktitle=gray,
  title=Readability Scoring Instruction,
  boxed title style={size=small,colframe=gray} ]
\small
\textbf{Scoring Instructions:} Please evaluate the charts based on the following criteria, with a score range from 1 to 5, where 1 indicates very poor quality and 5 indicates excellent quality. You should focus on the following aspects:

\vspace{0.5em}
\textbf{1. Chart Colors:}
\begin{itemize}
    \item Are the colors clear and natural, effectively conveying the information?
    \item Color blindness accessibility: Are the color combinations easy to distinguish, especially for users with color blindness?
\end{itemize}

\vspace{0.5em}
\textbf{2. Title and Axis Labels:}
\begin{itemize}
    \item Ensure the chart has a clear title.
    \item Do the X-axis and Y-axis labels exist, and are they complete?
    \item Check if the labels are difficult to read, e.g., are they written vertically instead of horizontally?
    \item The title should not be a direct question; instead, it should describe the data or trends being presented.
\end{itemize}

\vspace{0.5em}
\textbf{3. Legend Completeness:}
\begin{itemize}
    \item Is the legend complete, and does it clearly indicate the color labels for different data series?
    \item Ensure each color has a corresponding legend, making it easy for users to understand what the data represents.
\end{itemize}

\vspace{0.5em}
\textbf{Scoring Scale:}
\begin{itemize}
    \item \textbf{1 Point:} Very poor, unable to understand or severely lacking information.
    \item \textbf{2 Points:} Poor quality, multiple issues present, difficult to extract information.
    \item \textbf{3 Points:} Fair, conveys some information but still has room for improvement.
    \item \textbf{4 Points:} Good, generally clear charts with minor areas for improvement.
    \item \textbf{5 Points:} Excellent, outstanding chart design with clear and effective information presentation.
\end{itemize}

Please consider the above factors when assessing the charts and provide the appropriate score. Thank you for your cooperation and effort!
\end{tcolorbox}
\vspace{-7pt}
\caption{Instructions for human annorators in annotating readability scoring.}
\label{fig:scoring_instructions}
\vspace{1em}
\end{figure*}

\begin{figure*}[ht]
\centering
\vspace{1em}
\begin{tcolorbox}[enhanced,attach boxed title to top center={yshift=-3mm,yshifttext=-1mm},boxrule=0.9pt, 
  colback=gray!00,colframe=black!50,colbacktitle=gray,
  title=Visualization Comparison Guidance,
  boxed title style={size=small,colframe=gray} ]
\small
Welcome to the visualization comparison evaluation. Your task is to judge which model-generated visualization better meets the requirements of the natural language query.

\vspace{0.5em}
\textbf{Evaluation criteria:}
\begin{enumerate}
    \item \textbf{Appropriateness of chart type:} Check if the selected chart type is suitable for expressing the data and relationships required by the query.
    \item \textbf{Data completeness:} Ensure the chart includes all necessary data required by the query.
    \item \textbf{Readability:} Assess the clarity of the chart, accuracy of labels, and overall layout.
    \item \textbf{Aesthetics:} Consider if the chart's color scheme, proportions, and overall design are visually pleasing.
    \item \textbf{Information conveyance:} Judge if the chart effectively conveys the main information or insights required by the query.
\end{enumerate}

\vspace{0.5em}
\textbf{Evaluation process:}
\begin{enumerate}
    \item Carefully read the natural language query.
    \item Observe the visualization results generated by two models.
    \item Based on the above criteria, choose the better visualization or select a tie if they are equally good.
    \item If neither visualization satisfies the query requirements well, please choose the relatively better one.
\end{enumerate}

Remember, your evaluation will help us improve and compare different visualization models. Thank you for your participation!
\end{tcolorbox}
\vspace{-7pt}
\caption{Instructions for human annorators in visualization comparison.}
\label{fig:evaluation_instructions}
\vspace{1em}
\end{figure*}

\section{Additional Experiment Results}
\label{additional_experiment_result}

We also conducted a comparison experiment of different methods using matplotlib or seaborn library. Figure~\ref{fig: library} demonstrates the results, indicating that our method outperforms obviously other baselines not only with matplotlib but also seaborn.

In addition, we test techniques in the Validator Agent, such as Chain-of-Thought. As is shown in Table~\ref{tab:ablation_validator}, integrating Chain-of-Thought reasoning, may affect its performance badly, likely due to the simple refining task with complex reasoning. Moreover, using the original schema to check for false schema filtering seems to be useless in this case.

\section{Evaluation Results with Detailed Metrics}
We demonstrated the main results in Table~\ref{tab:performance_comparison}, and here we reported more detailed results of other metrics in Table~\ref{tab:detailed_results}, which underscored the error rates for each stage, including \textit{Invalid}, \textit{Illegal}, and \textit{Low Readability}. 

\begin{table*}[!ht]
\centering
\footnotesize
\scalebox{0.98}{
\begin{tabular}{ll|cc|cccc|cc}
\toprule[1.5pt]
\multirow{2}{*}{Method} & \multirow{2}{*}{Dataset} & \multicolumn{2}{c|}{Invalid} & \multicolumn{4}{c|}{Illegal} & \multicolumn{2}{c}{Low Readability} \\
&  & Execution & Surface. & Decon. & Chart Type & Data & Order & Layout & Scale\&Ticks \\
\midrule
\multicolumn{10}{c}{ \textbf{\textit{GPT-4o}}}\\
\midrule
\multirow{3}{*}{CoML4Vis} & All & 1.15 & 0.00 & 0.26 & 1.75 & 14.28 & 10.36 & 32.02 & 32.55 \\
& Single & 0.67 & 0.00 & 0.43 & 1.93 & 13.54 & 10.16 & 31.08 & 32.76 \\
& Multiple & 1.87 & 0.00 & 0.00 & 1.48 & 15.39 & 10.66 & 33.43 & 32.23 \\
\cmidrule{2-10}
\multirow{3}{*}{LIDA} & All & 6.61 & 0.00 & 1.60 & 3.24 & 40.53 & 4.07 & 32.68 & 15.77 \\
& Single & 1.13 & 0.00 & 2.11 & 0.89 & 12.26 & 6.79 & 53.93 & 26.22 \\
& Multiple & 14.80 & 0.00 & 0.79 & 8.51 & 80.53 & 0.00 & 1.24 & 0.21 \\
\cmidrule{2-10}
\multirow{3}{*}{Chat2Vis} & All & 16.05 & 0.00 & 0.62 & 3.99 & 30.14 & 5.96 & 2.37 & 20.88 \\
& Single & 0.86 & 0.00 & 0.75 & 2.30 & 10.78 & 9.73 & 3.97 & 34.63 \\
& Multiple & 38.74 & 0.00 & 0.43 & 6.51 & 59.08 & 0.32 & 0.00 & 0.34 \\
\cmidrule{2-10}
\multirow{3}{*}{nvAgent} & All & 0.97 & 0.00 & 0.08 & 1.28 & 11.07 & 4.05 & 5.07 & 40.03 \\
& Single & 0.72 & 0.00 & 0.14 & 1.27 & 9.88 & 3.60 & 3.92 & 39.36 \\
& Multiple & 1.34 & 0.00 & 0.00 & 1.30 & 12.84 & 4.73 & 6.79 & 41.03 \\
\midrule
\multicolumn{10}{c}{ \textbf{\textit{GPT-4o-mini}}}\\
\midrule
\multirow{3}{*}{CoML4Vis} & All & 4.23 & 0.00 & 0.20 & 2.31 & 16.64 & 11.83 & 35.23 & 29.35 \\
& Single & 0.36 & 0.00 & 0.26 & 2.32 & 13.80 & 11.67 & 35.92 & 32.22 \\
& Multiple & 10.01 & 0.00 & 0.10 & 2.31 & 20.87 & 12.07 & 34.19 & 25.05 \\
\cmidrule{2-10}
\multirow{3}{*}{LIDA} & All & 12.50 & 0.00 & 0.40 & 4.92 & 40.02 & 5.80 & 27.87 & 17.05 \\
& Single & 9.09 & 0.00 & 0.44 & 2.53 & 12.91 & 9.68 & 45.69 & 28.32 \\
& Multiple & 17.61 & 0.00 & 0.33 & 8.51 & 80.53 & 0.00 & 1.24 & 0.21 \\
\cmidrule{2-10}
\multirow{3}{*}{Chat2Vis} & All & 15.45 & 0.17 & 0.17 & 4.21 & 31.90 & 8.20 & 2.14 & 18.97 \\
& Single & 2.14 & 0.29 & 0.41 & 2.53 & 11.99 & 9.68 & 45.69 & 28.32 \\
& Multiple & 35.78 & 0.00 & 0.00 & 6.70 & 61.66 & 0.00 & 0.92 & 0.32 \\
\cmidrule{2-10}
\multirow{3}{*}{nvAgent} & All & 5.14 & 0.00 & 0.00 & 2.40 & 16.33 & 10.61 & 41.06 & 27.00 \\
& Single & 1.97 & 0.00 & 0.14 & 2.97 & 15.21 & 7.49 & 39.30 & 32.39 \\
& Multiple & 8.15 & 0.00 & 0.00 & 2.31 & 20.87 & 12.07 & 34.19 & 25.05 \\
\midrule
\multicolumn{10}{c}{ \textbf{\textit{GPT-3.5-turbo}}}\\
\midrule
\multirow{3}{*}{CoML4Vis} & All & 9.28 & 0.00 & 0.62 & 1.91 & 15.83 & 12.86 & 25.09 & 27.73 \\ 
& Single & 6.17 & 0.00 & 0.89 & 2.50 & 14.71 & 13.20 & 26.10 & 29.93 \\ 
& Multiple & 13.92 & 0.00 & 0.21 & 1.04 & 17.51 & 12.36 & 23.57 & 24.43 \\ 
\cmidrule{2-10} 
\multirow{3}{*}{LIDA} & All & 53.43 & 0.00 & 1.27 & 3.56 & 22.33 & 0.53 & 14.90 & 6.62 \\ 
& Single & 47.32 & 0.00 & 1.91 & 2.81 & 13.03 & 0.89 & 24.43 & 11.05 \\ 
& Multiple & 62.57 & 0.00 & 0.32 & 4.68 & 36.23 & 0.00 & 0.65 & 0.00 \\ 
\cmidrule{2-10} 
\multirow{3}{*}{Chat2Vis} & All & 18.68 & 0.00 & 0.28 & 3.66 & 32.47 & 7.20 & 25.45 & 20.15 \\ 
& Single & 3.90 & 0.00 & 0.47 & 2.78 & 15.62 & 12.01 & 41.74 & 33.38 \\ 
& Multiple & 40.77 & 0.00 & 0.00 & 4.97 & 57.66 & 0.00 & 1.12 & 0.37 \\ 
\cmidrule{2-10} 
\multirow{3}{*}{nvAgent} & All & 4.66 & 0.00 & 0.08 & 3.06 & 18.24 & 5.64 & 5.25 & 35.34 \\ 
& Single & 2.98 & 0.00 & 0.14 & 2.84 & 15.08 & 5.69 & 3.62 & 37.57 \\ 
& Multiple & 7.18 & 0.00 & 0.00 & 3.38 & 22.95 & 5.56 & 7.69 & 32.02 \\
\bottomrule[1.5pt]
\end{tabular}
}
\caption{Detailed error rates (\%) for different methods.} 
\label{tab:detailed_results}
\end{table*}

\section{Case Study}
\label{example}
\begin{figure*}[!ht]
    \centering
    \includegraphics[width=0.98\linewidth,scale=1.0]{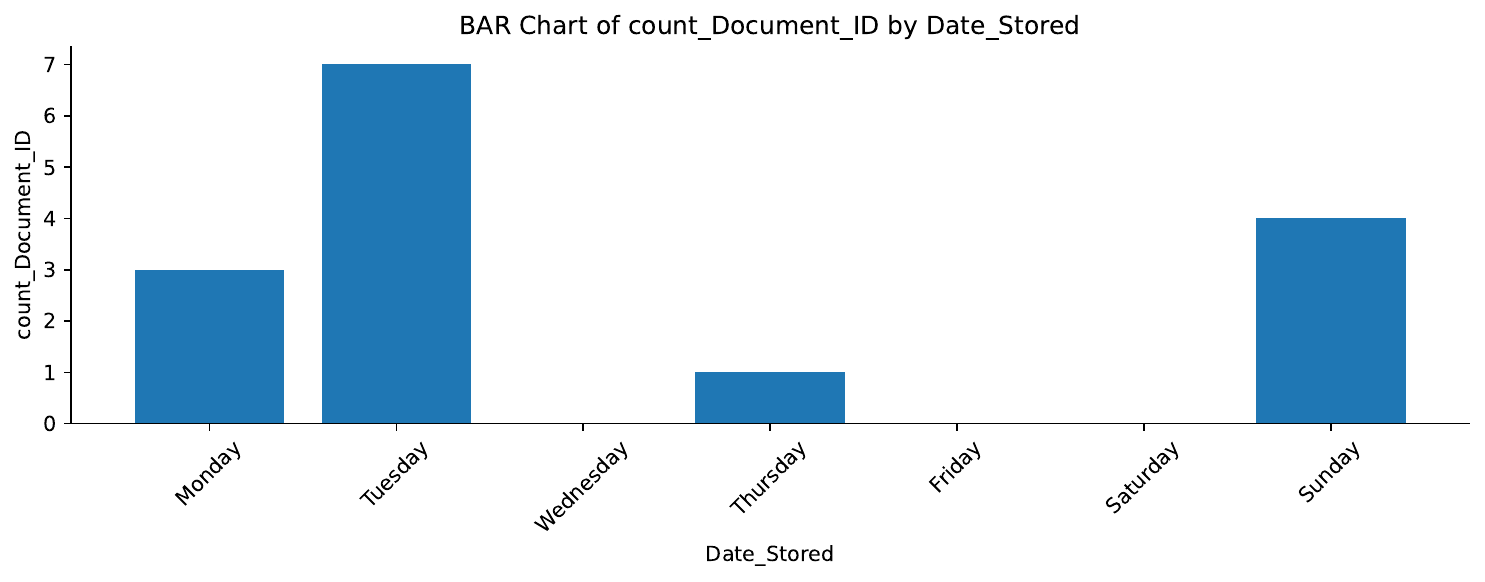}
    \caption{An example of generated bar chart using \system.}
    \label{fig:example_chart}
\end{figure*}

\begin{figure*}[htbp]
\centering
\begin{tcolorbox}[enhanced,attach boxed title to top center={yshift=-3mm,yshifttext=-1mm},boxrule=0.9pt, 
  colback=gray!00,colframe=black!50,colbacktitle=gray,
  title=Examples of \textsc{nvAgent} performance on different hardness levels,
  boxed title style={size=small,colframe=gray} ]
  
\textbf{Hardness Level:} Easy \\
\begin{minipage}{0.45\linewidth}
    \textbf{Dataset}: \textit{Single}\\
    \textbf{Input Tables}: basketball\_match\\
    \textbf{Input Query}: Show the relation between acc percent and all\_games\_percent for each ACC\_Home using a grouped scatter chart.\\
\end{minipage}\hfill
\begin{minipage}{0.45\linewidth}
    \centering
    \textbf{Response}:
    \includegraphics[width=\linewidth]{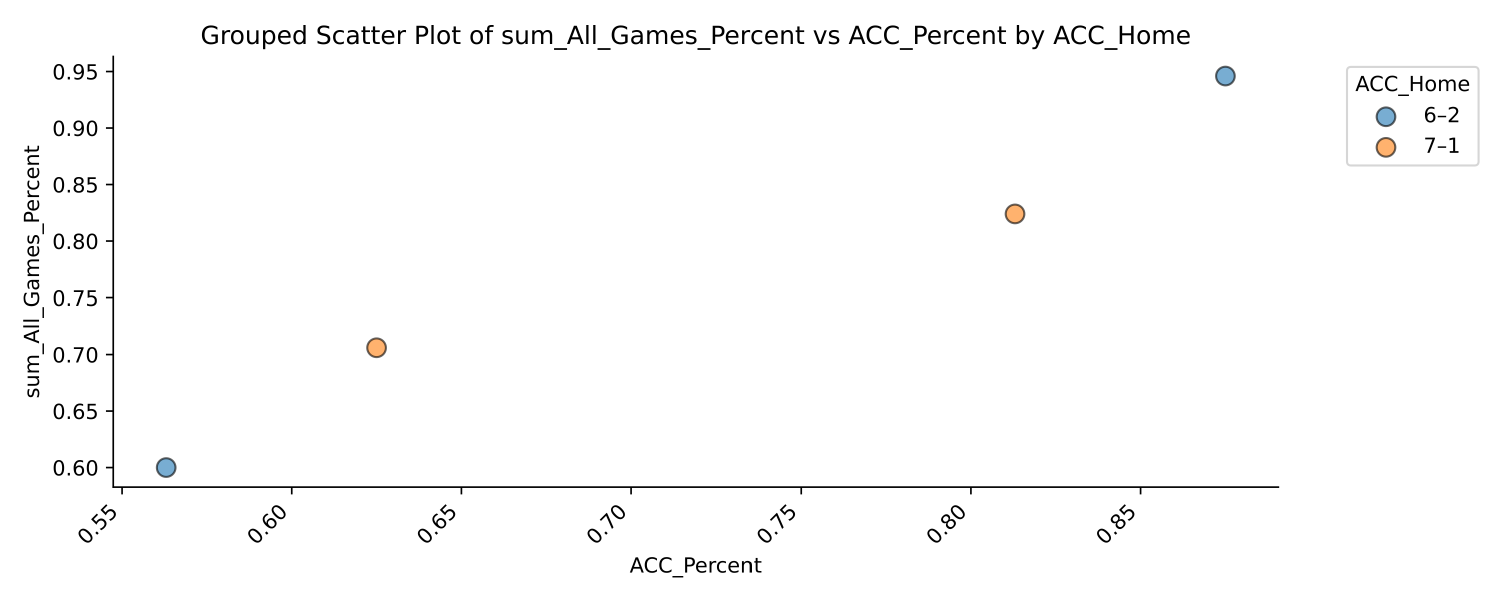} 
\end{minipage}
\tcbline

\textbf{Hardness Level:} Medium \\
\begin{minipage}{0.45\linewidth}
    \textbf{Dataset}: \textit{Multiple}\\
    \textbf{Input Tables}: car\_makers, car\_names, cars\_data, continents, countries, model\_list\\
    \textbf{Input Query}: Display a pie chart for what is the name of each continent and how many car makers are there in each one?\\
\end{minipage}\hfill
\begin{minipage}{0.55\linewidth}
    \centering
    \textbf{Response}:
    \includegraphics[width=\linewidth]{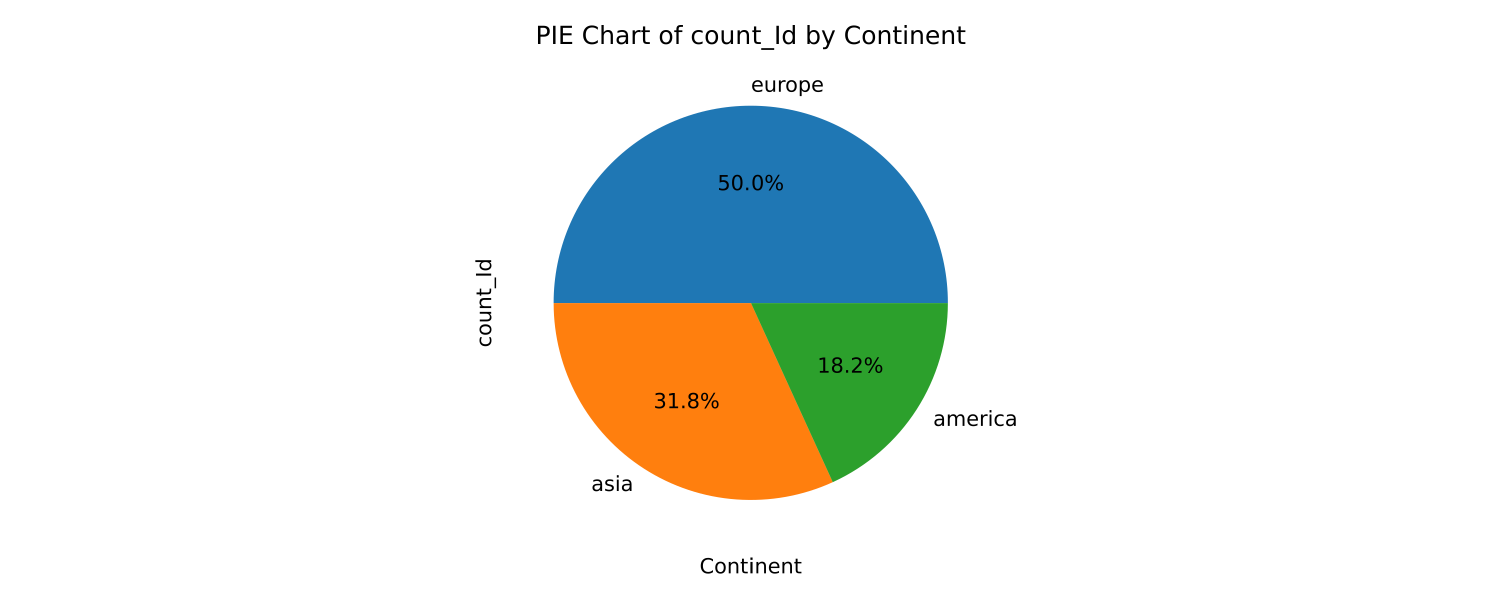} 
\end{minipage}
\tcbline

\textbf{Hardness Level:} Hard \\[1em]
\begin{minipage}{0.45\linewidth}
    \textbf{Dataset}: \textit{Multiple}\\
    \textbf{Input Tables}: advisor, classroom, course, department, instructor, prereq, section, student, takes, teaches, time\_slot\\
    \textbf{Input Query}: Find the number of courses offered by Psychology department in each year with a line chart.\\
\end{minipage}\hfill
\begin{minipage}{0.45\linewidth}
    \centering
    \textbf{Response}:
    \includegraphics[width=\linewidth]{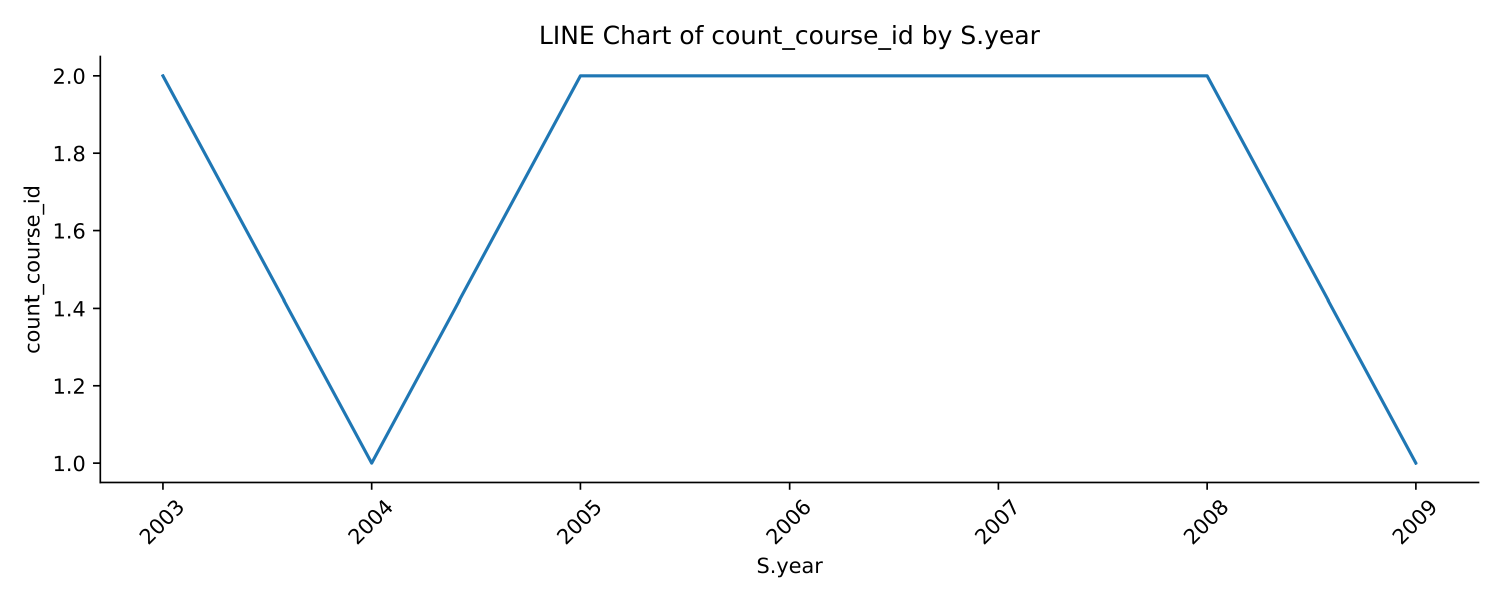} 
\end{minipage}
\tcbline

\textbf{Hardness Level:} Extra Hard \\[1em]
\begin{minipage}{0.45\linewidth}
    \textbf{Dataset}: \textit{Multiple}\\
    \textbf{Input Tables}: Accounts, Documents, Documents\_with\_Expenses, Projects, Ref- \_Budget\_Codes, Ref\_Document\_Types, Statements\\
    \textbf{Input Query}: How many documents are created in each day? Bin the document date by weekday and group by document type description with a stacked bar chart, I want to sort Y in desc order.\\
\end{minipage}\hfill
\begin{minipage}{0.45\linewidth}
    \centering
    \textbf{Response}:
    \includegraphics[width=\linewidth]{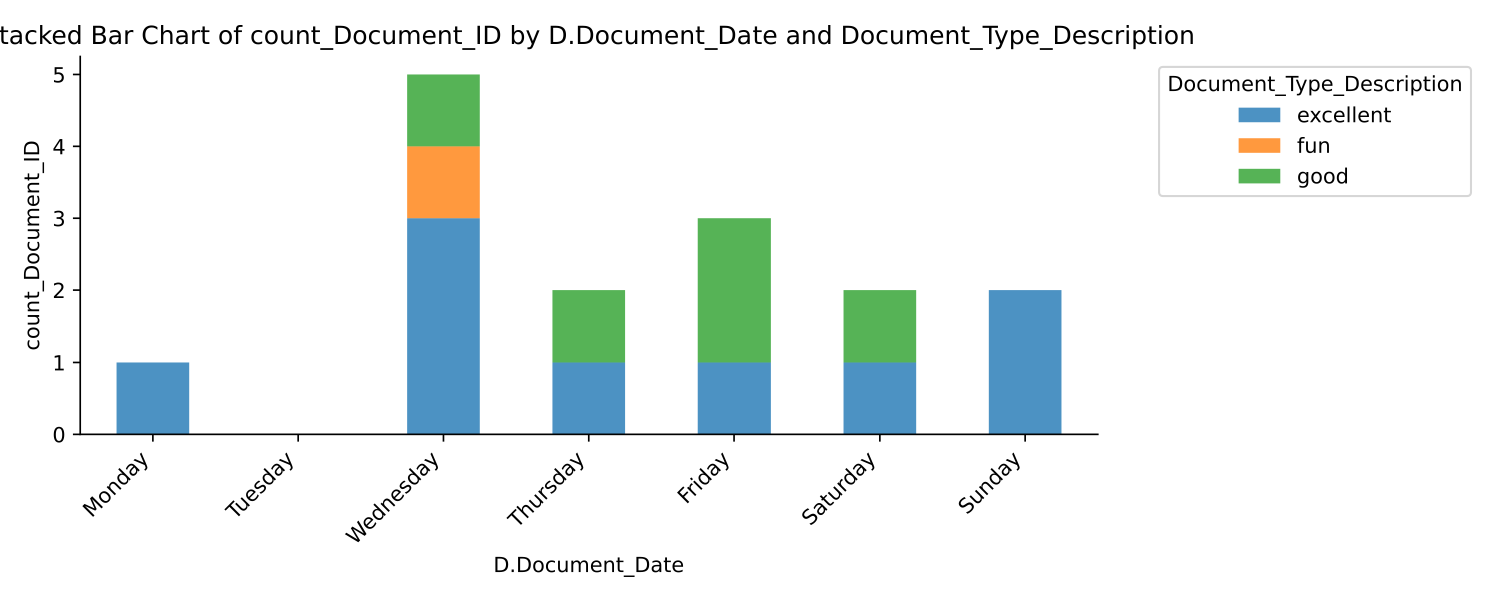} 
\end{minipage}

\end{tcolorbox}
    \caption{Examples of \textsc{nvAgent}'s performance on different hardness levels in VisEval (easy, medium, hard, and extra hard.}
    \label{hardness case}
\end{figure*}

Figure~\ref{fig:nl_vql} shows an example of a natural language query with its corresponding VQL representation. The output Python code for visualization and the final bar chart are demonstrated in Figure~\ref{python code} and Figure~\ref{fig:example_chart}, respectively.
Furthermore, we provide a case study of \system performance on four hardness-level NL2Vis problems in VisEval in Figure \ref{hardness case}.

The easy case demonstrates accurate grouping in scatter plot relationships. The medium case shows correct handling of multi-table joins for continent-wise statistics. The hard case exhibits temporal data visualization with proper filtering. The extra hard case showcases complex operations including weekday binning and stacked visualization. These cases highlight our system's consistent performance across varying task complexities, particularly excelling in multiple table scenarios and complex aggregations.

\begin{figure*}[htbp]
\centering
\begin{tcolorbox}[enhanced,attach boxed title to top center={yshift=-3mm,yshifttext=-1mm},boxrule=0.9pt, 
  colback=gray!00,colframe=black!50,colbacktitle=gray,
  title=An Example of Natural Language Query and  Corresponding VQL,,
  boxed title style={size=small,colframe=gray} ]

\textbf{Natural Language Query:}\\
How many documents are stored? Bin the store date by weekday in a bar chart.\\
\tcbline
\textbf{Corresponding VQL:}\\
Visualize BAR \\
SELECT Date\_Stored, COUNT(Document\_ID)\\
FROM All\_Documents \\
GROUP BY Date\_Stored \\
BIN Date\_Stored BY WEEKDAY\\
\end{tcolorbox}
\caption{The natural language query case and its corresponding output VQL representation.}
\label{fig:nl_vql}
\end{figure*}

\lstset{
    basicstyle=\ttfamily\small,
    breaklines=true,
    numbers=left,
    numberstyle=\tiny,
    frame=single,
    showstringspaces=false,
    tabsize=4,
    keywordstyle=\color{blue},
    commentstyle=\color{green!60!black},
    stringstyle=\color{purple},
    breakatwhitespace=false,
    keepspaces=true
}
\begin{figure*}[htbp]
\begin{lstlisting}[
    language=Python,
    frame=single,
    numbers=left,
    numberstyle=\tiny,
    basicstyle=\tiny\ttfamily,    
    breaklines=true,
    breakatwhitespace=false,
    showstringspaces=false,
    framesep=1mm,                 % 减小框架边距
    framexleftmargin=1mm,         % 减小框架左边距
    columns=flexible,             % 灵活列宽
    keepspaces=true
]
import matplotlib.pyplot as plt 
import pandas as pd 
import os 
import duckdb

# Set data folder path
data_folder = 'E:/visEval_dataset/databases/cre_Doc_Tracking_DB'  

# Connect to database 
con = duckdb.connect(database=':memory:')  

# Read all CSV files and create view
csv_files = [f for f in os.listdir(data_folder) if f.endswith('.csv')] 
for file in csv_files:
    table_name = os.path.splitext(file)[0]
    con.execute(f"CREATE VIEW {table_name} AS SELECT * FROM read_csv_auto('{os.path.join(data_folder, file)}')")  
    
# Execute SQL query 
sql = f''' 
    SELECT Date_Stored, COUNT(Document_ID) AS count_Document_ID 
    FROM All_Documents 
    GROUP BY Date_Stored 
''' 
df = con.execute(sql).fetchdf() 
con.close()  

# Rename columns 
df.columns = ['Date_Stored','count_Document_ID'] 

# Apply binning operation
flag = True 
df['Date_Stored'] = pd.to_datetime(df['Date_Stored']) 
df['Date_Stored'] = df['Date_Stored'].dt.day_name()  

# Group by and calculate count 
if flag:
    df = df.groupby('Date_Stored').sum().reset_index() 

# Ensure all seven days of the week are included 
weekday_order = ['Monday', 'Tuesday', 'Wednesday', 'Thursday', 
                 'Friday', 'Saturday', 'Sunday'] 
df = df.set_index('Date_Stored').reindex(weekday_order, fill_value=0).reset_index()
df['Date_Stored'] = pd.Categorical(df['Date_Stored'], 
                                  categories=weekday_order, ordered=True) 
df = df.sort_values('Date_Stored')

# Create visualization 
fig, ax = plt.subplots(1, 1, figsize=(10, 4)) 
ax.spines['top'].set_visible(False) 
ax.spines['right'].set_visible(False) 
ax.bar(df['Date_Stored'], df['count_Document_ID']) 
ax.set_xlabel('Date_Stored') 
ax.set_ylabel('count_Document_ID') 
ax.set_title(f'BAR Chart of count_Document_ID by Date_Stored') 
plt.xticks(rotation=45) 
plt.tight_layout()  
plt.show()
\end{lstlisting}
\caption{An example of python code generating module within \system.}
\label{python code}
\end{figure*}

\clearpage
\onecolumn
\section{Prompts Details}
\label{prompt_details}
We provide detailed prompt design of our \system as follows.

\begin{promptbox}[Prompt template for Processor Agent]
You are an experienced and professional database administrator. Given a database schema and a user query, your task is to analyze the query, filter the relevant schema, generate an optimized representation, and classify the query difficulty. \\
\\
Now you can think step by step, following these instructions below. \\
\textbf{[Instructions]} \\
1. Schema Filtering: \\
\text{\ \ \ \ }- Identify the tables and columns that are relevant to the user query.\\
\text{\ \ \ \ }- Only exclude columns that are completely irrelevant.\\
\text{\ \ \ \ }- The output should be \{\{tables: [columns]\}\}.\\
\text{\ \ \ \ }- Keep the columns needed to be primary keys and foreign keys in the filtered schema.\\
\text{\ \ \ \ }- Keep the columns that seem to be similar with other columns of another table.\\
\\
2. New Schema Generation:\\
\text{\ \ \ \ }- Generate a new schema of the filtered schema, based on the given database schema and your filtered schema.\\
\\
3. Augmented Explanation:\\
\text{\ \ \ \ }- Provide a concise summary of the filtered schema to give additional knowledge.\\
\text{\ \ \ \ }- Include the number of tables, total columns, and any notable relationships or patterns.\\
\\
4. Classification:\\
For the database new schema, classify it as SINGLE or MULTIPLE based on the tables number.\\
\text{\ \ \ \ }- if tables number >= 2: predict MULTIPLE\\
\text{\ \ \ \ }- elif only one table: predict SINGLE\\
\\
==============================\\
Here is a typical example:\\
\textbf{[Database Schema]}\\
\textbf{[DB\_ID]} dorm\_1\\
\textbf{[Schema]}\\
\# Table: Student\\
\text{[}\\
  \text{\ \ \ \ }(stuid, And This is a id type column),\\
  \text{\ \ \ \ }(lname, Value examples: [`Smith', `Pang', `Lee', `Adams', `Nelson', `Wilson'].),\\
  \text{\ \ \ \ }(fname, Value examples: [`Eric', `Lisa', `David', `Sarah', `Paul', `Michael'].),\\
  \text{\ \ \ \ }(age, Value examples: [18, 20, 17, 19, 21, 22].),\\
  \text{\ \ \ \ }(sex, Value examples: [`M', `F'].),\\
  \text{\ \ \ \ }(major, Value examples: [600, 520, 550, 50, 540, 100].),\\
  \text{\ \ \ \ }(advisor, And this is a number type column),\\
  \text{\ \ \ \ }(city code, Value examples: [`PIT', `BAL', `NYC', `WAS', `HKG', `PHL'].)\\
\text{]}\\
\# Table: Dorm\\
\text{[}\\
  \text{\ \ \ \ }(dormid, And This is a id type column),\\
  \text{\ \ \ \ }(dorm name, Value examples: [`Anonymous Donor Hall', `Bud Jones Hall', `Dorm-plex 2000', `Fawlty Towers', `Grad Student Asylum', `Smith Hall'].),\\
  \text{\ \ \ \ }(student capacity, Value examples: [40, 85, 116, 128, 256, 355].),
  (gender, Value examples: [`X', `F', `M'].)\\
\text{]}\\
\# Table: Dorm\_amenity\\
\text{[}\\
  \text{\ \ \ \ }(amenid, And This is a id type column),\\
  \text{\ \ \ \ }(amenity name, Value examples: [`4 Walls', `Air Conditioning', `Allows Pets', `Carpeted Rooms', `Ethernet Ports', `Heat'].)\\
\text{]}\\
\# Table: Has\_amenity\\
\text{[}\\
  \text{\ \ \ \ }(dormid, And This is a id type column),\\
  \text{\ \ \ \ }(amenid, And This is a id type column)\\
\text{]}\\
\# Table: Lives\_in\\
\text{[}\\
  \text{\ \ \ \ }(stuid, And This is a id type column),\\
  \text{\ \ \ \ }(dormid, And This is a id type column),\\
  \text{\ \ \ \ }(room number, And this is a number type column)\\
\text{]}\\
\\
\textbf{[Query]}\\
Find the first name of students who are living in the Smith Hall, and count them by a pie chart\\
\\
Now we can think step by step\\
\textbf{[Filtered Schema]}\\
\{\\
  "Student": ["stuid", "fname"],\\
  "Dorm": ["dormid", "dorm name"],\\
  "Lives\_in": ["stuid", "dormid"]\\
\}\\
\\
\textbf{[New Schema]}\\
\# Table: Student\\
\text{[}\\
  (stuid, And This is a id type column),\\
  (fname, Value examples: [`Eric', `Lisa', `David', `Sarah', `Paul', `Michael'].),\\
\text{]}\\
\# Table: Dorm
\text{[}\\
  (dormid, And This is a id type column),\\
  (dorm name, Value examples: [`Anonymous Donor Hall', `Bud Jones Hall', `Dorm-plex 2000', `Fawlty Towers', `Grad Student Asylum', `Smith Hall'].),\\
\text{]}\\
\# Table: Lives\_in\\
\text{[}\\
  (stuid, And This is a id type column),\\
  (dormid, And This is a id type column),\\
\text{]}\\
\textbf{[Augmented Explanation]}\\
The filtered schema consists of 3 tables (Student, Dorm, and Lives\_in) with a total of 6 relevant columns. There is a many-to-one relationship between Student and Dorm through the Lives\_in junction table. The query involves joining these three tables to find students living in a specific dorm (Smith Hall).\\
\\
Key points:\\
1. The Lives\_in table acts as a bridge between Student and Dorm, allowing for the association of students with their dorms.\\
2. The `dorm name' column in the Dorm table is crucial for filtering the specific dorm (Smith Hall).\\
3. The `fname' column from the Student table is required for the final output.\\
\\
\textbf{[Classification]}\\
MULTIPLE\\
\\
==============================\\
Here is a new question:\\
\\
\textbf{[DB\_ID]} \{db\_id\}\\
\textbf{[Database Schema]}\\
\{db\_schema\}\\
\\
\textbf{[Query]}\\
\{query\}\\
\\
Now give your answer following this format strictly without other explanation:\\
\\
\textbf{[Filtered Schema]}\\
\\
\textbf{[New Schema]}\\
\\
\textbf{[Augmented Explanation]}\\
\\
\textbf{[Classification]}\\
\\
\end{promptbox}

\begin{promptbox}[Prompt template for multiple classification]
Given a [Database schema] with [Augmented Explanation] and a [Question], generate a valid VQL (Visualization Query Language) sentence. VQL is similar to SQL but includes visualization components. \\
\\
Now you can think step by step, following these instructions below. \\
\textbf{[Background]} \\
VQL Structure:\\
Visualize [TYPE] SELECT [COLUMNS] FROM [TABLES] [JOIN] [WHERE] [GROUP BY] [ORDER BY] [BIN BY]\\
\\
You can consider a VQL sentence as "VIS TYPE + SQL + BINNING"\\
You must consider which part in the sketch is necessary, which is unnecessary, and construct a specific sketch for the natural language query.\\
\\
Key Components:\\
1. Visualization Type: bar, pie, line, scatter, stacked bar, grouped line, grouped scatter\\
2. SQL Components: SELECT, FROM, JOIN, WHERE, GROUP BY, ORDER BY\\
3. Binning: BIN [COLUMN] BY [INTERVAL], [INTERVAL]: [YEAR, MONTH, DAY, WEEKDAY]\\
\\
When generating VQL, we should always consider special rules and constraints:\\
\textbf{[Special Rules]} \\
a. For simple visualizations:\\
    \text{\ \ \ \ }- SELECT exactly TWO columns, X-axis and Y-axis(usually aggregate function)\\
b. For complex visualizations (STACKED BAR, GROUPED LINE, GROUPED SCATTER):\\
    \text{\ \ \ \ }- SELECT exactly THREE columns in this order!!!:\\
        \text{\ \ \ \ }\text{\ \ \ \ }1. X-axis\\
        \text{\ \ \ \ }\text{\ \ \ \ }2. Y-axis (aggregate function)\\
        \text{\ \ \ \ }\text{\ \ \ \ }3. Grouping column\\
c. When "COLORED BY" is mentioned in the question:\\
    \text{\ \ \ \ }- Use complex visualization type(STACKED BAR for bar charts, GROUPED LINE for line charts, GROUPED SCATTER for scatter charts)\\
    \text{\ \ \ \ }- Make the "COLORED BY" column the third SELECT column\\
    \text{\ \ \ \ }- Do NOT include "COLORED BY" in the final VQL\\     
d. Aggregate Functions:\\
    \text{\ \ \ \ }- Use COUNT for counting occurrences\\
    \text{\ \ \ \ }- Use SUM only for numeric columns\\
    \text{\ \ \ \ }- When in doubt, prefer COUNT over SUM\\
e. Time based questions:\\
    \text{\ \ \ \ }- Always use BIN BY clause at the end of VQL sentence\\
    \text{\ \ \ \ }- When you meet the questions including "year", "month", "day", "weekday"\\
    \text{\ \ \ \ }- Avoid using window function, just use BIN BY to deal with time base queries\\
\textbf{[Constraints]} \\
- In SELECT <column>, make sure there are at least two selected!!!\\
- In FROM <table> or JOIN <table>, do not include unnecessary table\\
- Use only table names and column names from the given database schema\\
- Enclose string literals in single quotes\\
- If [Value examples] of <column> has `None' or None, use JOIN <table> or WHERE <column> is NOT NULL is better\\
- Ensure GROUP BY precedes ORDER BY for distinct values\\
- NEVER use window functions in SQL\\
\\
Now we could think step by step:\\
1. First choose visualize type and binning, then construct a specific sketch for the natural language query\\
2. Second generate SQL components following the sketch.\\
3. Third add Visualize type and BINNING into the SQL components to generate final VQL\\
\\
==============================\\
Here is a typical example:\\
\textbf{[Database Schema]}\\
\# Table: Orders, (orders)\\
\text{[}\\
  \text{\ \ \ \ }(order\_id, order id, And this is a id type column),\\
  \text{\ \ \ \ }(customer\_id, customer id, And this is a id type column),\\
  \text{\ \ \ \ }(order\_date, order date, Value examples: [`2023-01-15', `2023-02-20', `2023-03-10'].),\\
  \text{\ \ \ \ }(total\_amount, total amount, Value examples: [100.00, 200.00, 300.00, 400.00, 500.00].)\\
\text{]}\\
\# Table: Customers, (customers)\\
\text{[}\\
  \text{\ \ \ \ }(customer\_id, customer id, And this is a id type column),\\
  \text{\ \ \ \ }(customer\_name, customer name, Value examples: [`John', `Emma', `Michael', `Sophia', `William'].),\\
  \text{\ \ \ \ }(customer\_type, customer type, Value examples: [`Regular', `VIP', `New'].)\\
\text{]}\\
\textbf{[Augmented Explanation]}\\
The filtered schema consists of 2 tables (Orders and Customers) with a total of 7 relevant columns. There is a one-to-many relationship between Customers and Orders through the customer\_id foreign key.\\
\\
Key points:\\
1. The Orders table contains information about individual orders, including the order date and total amount.\\
2. The Customers table contains customer information, including their name and type (Regular, VIP, or New).\\
3. The customer\_id column links the two tables, allowing us to associate orders with specific customers.\\
4. The order\_date column in the Orders table will be used for monthly grouping and binning.\\
5. The total\_amount column in the Orders table needs to be summed for each group.\\
6. The customer\_type column in the Customers table will be used for further grouping and as the third dimension in the stacked bar chart.\\
\\

The query involves joining these two tables to analyze order amounts by customer type and month, which requires aggregation and time-based binning.\\
\\
\textbf{[Question]}\\
Show the total order amount for each customer type by month in a stacked bar chart.\\
\\
Decompose the task into sub tasks, considering [Background] [Special Rules] [Constraints], and generate the VQL after thinking step by step:\\
\\
\textbf{Sub task 1:} First choose visualize type and binning, then construct a specific sketch for the natural language query\\
Visualize type: STACKED BAR, BINNING: True\\
VQL Sketch:\\
Visualize STACKED BAR SELECT \_ , \_ , \_ FROM \_ JOIN \_ ON \_ GROUP BY \_ BIN \_ BY MONTH\\
\\
\textbf{Sub task 2:} Second generate SQL components following the sketch.\\
Let's think step by step:\\
1. We need to select 3 columns for STACKED BAR chart, order\_date as X-axis, SUM(total\_amout) as Y-axis, customer\_type as group column.\\
2. We need to join the Orders and Customers tables.\\
3. We need to group by customer type.\\
4. We do not need to use any window function for MONTH.\\
\\
\text{sql}\\
```sql\\
SELECT O.order\_date, SUM(O.total\_amount), C.customer\_type\\
FROM Orders AS O\\
JOIN Customers AS C ON O.customer\_id = C.customer\_id\\
GROUP BY C.customer\_type\\
```\\
\\
\textbf{Sub task 3:} Third add Visualize type and BINNING into the SQL components to generate final VQL\\
\textbf{Final VQL:}\\
Visualize STACKED BAR SELECT O.order\_date, SUM(O.total\_amount), C.customer\_type FROM Orders O JOIN Customers C ON O.customer\_id = C.customer\_id GROUP BY C.customer\_type BIN O.order\_date BY MONTH\\
\\
==============================\\
Here is a new question:\\
\\
\textbf{[Database Schema]}\\
\{desc\_str\}\\
\\
\textbf{[Augmented Explanation]}\\
\{augmented\_explanation\}\\
\\
\textbf{[Query]}\\
\{query\}\\
\\
Now, please generate a VQL sentence for the database schema and question after thinking step by step.\\

\end{promptbox}

\begin{promptbox}[Prompt template for single classification]
Given a [Database schema] with [Augmented Explanation] and a [Question], generate a valid VQL (Visualization Query Language) sentence. VQL is similar to SQL but includes visualization components. \\
\\
Now you can think step by step, following these instructions below. \\
\textbf{[Background]} \\
VQL Structure:\\
Visualize [TYPE] SELECT [COLUMNS] FROM [TABLES] [JOIN] [WHERE] [GROUP BY] [ORDER BY] [BIN BY]\\
\\
You can consider a VQL sentence as "VIS TYPE + SQL + BINNING"\\
You must consider which part in the sketch is necessary, which is unnecessary, and construct a specific sketch for the natural language query.\\
\\
Key Components:\\
1. Visualization Type: bar, pie, line, scatter, stacked bar, grouped line, grouped scatter\\
2. SQL Components: SELECT, FROM, JOIN, WHERE, GROUP BY, ORDER BY\\
3. Binning: BIN [COLUMN] BY [INTERVAL], [INTERVAL]: [YEAR, MONTH, DAY, WEEKDAY]\\
\\
When generating VQL, we should always consider special rules and constraints:\\
\textbf{[Special Rules]} \\
a. For simple visualizations:\\
    \text{\ \ \ \ }- SELECT exactly TWO columns, X-axis and Y-axis(usually aggregate function)\\
b. For complex visualizations (STACKED BAR, GROUPED LINE, GROUPED SCATTER):\\
    \text{\ \ \ \ }- SELECT exactly THREE columns in this order!!!:\\
        \text{\ \ \ \ }\text{\ \ \ \ }1. X-axis\\
        \text{\ \ \ \ }\text{\ \ \ \ }2. Y-axis (aggregate function)\\
        \text{\ \ \ \ }\text{\ \ \ \ }3. Grouping column\\
c. When "COLORED BY" is mentioned in the question:\\
    \text{\ \ \ \ }- Use complex visualization type(STACKED BAR for bar charts, GROUPED LINE for line charts, GROUPED SCATTER for scatter charts)\\
    \text{\ \ \ \ }- Make the "COLORED BY" column the third SELECT column\\
    \text{\ \ \ \ }- Do NOT include "COLORED BY" in the final VQL\\     
d. Aggregate Functions:\\
    \text{\ \ \ \ }- Use COUNT for counting occurrences\\
    \text{\ \ \ \ }- Use SUM only for numeric columns\\
    \text{\ \ \ \ }- When in doubt, prefer COUNT over SUM\\
e. Time based questions:\\
    \text{\ \ \ \ }- Always use BIN BY clause at the end of VQL sentence\\
    \text{\ \ \ \ }- When you meet the questions including "year", "month", "day", "weekday"\\
    \text{\ \ \ \ }- Avoid using window function, just use BIN BY to deal with time base queries\\
\textbf{[Constraints]} \\
- In SELECT <column>, make sure there are at least two selected!!!\\
- In FROM <table> or JOIN <table>, do not include unnecessary table\\
- Use only table names and column names from the given database schema\\
- Enclose string literals in single quotes\\
- If [Value examples] of <column> has `None' or None, use JOIN <table> or WHERE <column> is NOT NULL is better\\
- Ensure GROUP BY precedes ORDER BY for distinct values\\
- NEVER use window functions in SQL\\
\\
Now we could think step by step:\\
1. First choose visualize type and binning, then construct a specific sketch for the natural language query\\
2. Second generate SQL components following the sketch.\\
3. Third add Visualize type and BINNING into the SQL components to generate final VQL\\
\\
==============================\\
Here is a typical example:\\
\textbf{[Database Schema]}\\
\# Table: course, (course)\\
\text{[}\\
  \text{\ \ \ \ }(course\_id, course id, Value examples: [101, 696, 656, 659]. And this is an id type column),\\
  \text{\ \ \ \ }(title, title, Value examples: [`Geology', `Differential Geometry', `Compiler Design', `International Trade', `Composition and Literature', `Environmental Law'].),\\
  \text{\ \ \ \ }(dept\_name, dept name, Value examples: [`Cybernetics', `Finance', `Psychology', `Accounting', `Mech. Eng.', `Physics'].),\\
  \text{\ \ \ \ }(credits, credits, Value examples: [3, 4].)\\
\text{]}\\
\# Table: section, (section)\\
\text{[}\\
  \text{\ \ \ \ }(course\_id, course id, Value examples: [362, 105, 960, 468]. And this is an id type column),\\
  \text{\ \ \ \ }(sec\_id, sec id, Value examples: [1, 2, 3]. And this is an id type column),\\
  \text{\ \ \ \ }(semester, semester, Value examples: [`Fall', `Spring'].),\\
  \text{\ \ \ \ }(year, year, Value examples: [2002, 2006, 2003, 2007, 2010, 2008].),\\
  \text{\ \ \ \ }(building, building, Value examples: [`Saucon', `Taylor', `Lamberton', `Power', `Fairchild', `Main'].),\\
  \text{\ \ \ \ }(room\_number, room number, Value examples: [180, 183, 134, 143].),\\
  \text{\ \ \ \ }(time\_slot\_id, time slot id, Value examples: [`D', `J', `M', `C', `E', `F']. And this is an id type column)\\
\text{]}\\
\textbf{[Augmented Explanation]}\\
The filtered schema consists of 2 tables (course and section) with a total of 11 relevant columns. There is a one-to-many relationship between course and section through the course\_id foreign key.\\
\\
Key points:\\
1. The course table contains information about individual courses, including the course title, department, and credits.\\
2. The section table contains information about specific sections of courses, including the semester, year, building, room number, and time slot.\\
3. The course\_id column links the two tables, allowing us to associate sections with specific courses.\\
4. The dept\_name column in the course table will be used to filter for Psychology department courses.\\
5. The year column in the section table will be used for yearly grouping and binning.\\
6. We need to count the number of courses offered each year, which requires aggregation and time-based binning.\\
\\
The query involves joining these two tables to analyze the number of courses offered by the Psychology department each year, which requires aggregation and time-based binning.\\
\\
\textbf{[Question]}\\
Find the number of courses offered by Psychology department in each year with a line chart.\\
\\
Decompose the task into sub tasks, considering [Background] [Special Rules] [Constraints], and generate the VQL after thinking step by step:\\
\\
\textbf{Sub task 1:} First choose visualize type and binning, then construct a specific sketch for the natural language query\\
Visualize type: LINE, BINNING: True\\
VQL Sketch:\\
Visualize LINE SELECT \_ , \_ FROM \_ JOIN \_ ON \_ WHERE \_ BIN \_ BY YEAR\\
\\
\textbf{Sub task 2:} Second generate SQL components following the sketch.\\
Let's think step by step:\\
1. We need to select 2 columns for LINE chart, year as X-axis, COUNT(year) as Y-axis.\\
2. We need to join the course and section tables to get the number of courses offered by the Psychology department in each year.\\
3. We need to filter the courses by the Psychology department.\\
4. We do not need to use any window function for YEAR.\\
\\
\text{sql}\\
```sql\\
SELECT S.year, COUNT(S.year)\\
FROM course AS C\\
JOIN section AS S ON C.course\_id = S.course\_id\\
WHERE C.dept\_name = `Psychology'\\
```\\
\\
\textbf{Sub task 3:} Third add Visualize type and BINNING into the SQL components to generate final VQL\\
\textbf{Final VQL:}\\
Visualize LINE SELECT S.year, COUNT(S.year) FROM course C JOIN section S ON C.course\_id = S.course\_id WHERE C.dept\_name = `Psychology' BIN S.year BY YEAR\\
\\
==============================\\
Here is a new question:\\
\\
\textbf{[Database Schema]}\\
\{desc\_str\}\\
\\
\textbf{[Augmented Explanation]}\\
\{augmented\_explanation\}\\
\\
\textbf{[Query]}\\
\{query\}\\
\\
Now, please generate a VQL sentence for the database schema and question after thinking step by step.\\

\end{promptbox}

\begin{promptbox}[Prompt template for Validator Agent]
As an AI assistant specializing in data visualization and VQL (Visualization Query Language), your task is to refine a VQL query that has resulted in an error. Please approach this task systematically, thinking step by step.\\
\textbf{[Background]}\\
VQL Structure:\\
Visualize [TYPE] SELECT [COLUMNS] FROM [TABLES] [JOIN] [WHERE] [GROUP BY] [ORDER BY] [BIN BY]\\
\\
You can consider a VQL sentence as "VIS TYPE + SQL + BINNING"\\
\\
Key Components:\\
1. Visualization Type: bar, pie, line, scatter, stacked bar, grouped line, grouped scatter\\
2. SQL Components: SELECT, FROM, JOIN, WHERE, GROUP BY, ORDER BY\\
3. Binning: BIN [COLUMN] BY [INTERVAL], [INTERVAL]: [YEAR, MONTH, DAY, WEEKDAY]\\
\\
When refining VQL, we should always consider special rules and constraints:\\
\textbf{[Special Rules]} \\
a. For simple visualizations:\\
    \text{\ \ \ \ }- SELECT exactly TWO columns, X-axis and Y-axis(usually aggregate function)\\
b. For complex visualizations (STACKED BAR, GROUPED LINE, GROUPED SCATTER):\\
    \text{\ \ \ \ }- SELECT exactly THREE columns in this order!!!:\\
        \text{\ \ \ \ }\text{\ \ \ \ }1. X-axis\\
        \text{\ \ \ \ }\text{\ \ \ \ }2. Y-axis (aggregate function)\\
        \text{\ \ \ \ }\text{\ \ \ \ }3. Grouping column\\
c. When "COLORED BY" is mentioned in the question:\\
    \text{\ \ \ \ }- Use complex visualization type(STACKED BAR for bar charts, GROUPED LINE for line charts, GROUPED SCATTER for scatter charts)\\
    \text{\ \ \ \ }- Make the "COLORED BY" column the third SELECT column\\
    \text{\ \ \ \ }- Do NOT include "COLORED BY" in the final VQL\\     
d. Aggregate Functions:\\
    \text{\ \ \ \ }- Use COUNT for counting occurrences\\
    \text{\ \ \ \ }- Use SUM only for numeric columns\\
    \text{\ \ \ \ }- When in doubt, prefer COUNT over SUM

e. Time based questions:\\
    \text{\ \ \ \ }- Always use BIN BY clause at the end of VQL sentence\\
    \text{\ \ \ \ }- When you meet the questions including "year", "month", "day", "weekday"\\
    \text{\ \ \ \ }- Avoid using time function, just use BIN BY to deal with time base queries\\
\\
\textbf{[Constraints]} \\
- In FROM <table> or JOIN <table>, do not include unnecessary table\\
- Use only table names and column names from the given database schema\\
- Enclose string literals in single quotes\\
- If [Value examples] of <column> has `None' or None, use JOIN <table> or WHERE <column> is NOT NULL is better\\
- ENSURE GROUP BY clause cannot contain aggregates\\
- NEVER use date functions in SQL\\
\\
\textbf{[Query]} \\
\{query\}\\
\\
\textbf{[Database info]} \\
\{db\_info\}\\
\\
\textbf{[Current VQL]} \\
\{vql\}\\
\\
\textbf{[Error]} \\
\{error\}\\
\\
Now, please analyze and refine the VQL, please provide:\\
\\
\textbf{[Explanation]}\\
\text{[}Provide a detailed explanation of your analysis process, the issues identified, and the changes made. Reference specific steps where relevant.\text{]}\\
\\
\textbf{[Corrected VQL]}\\
\text{[}Present your corrected VQL here. Ensure it's on a single line without any line breaks.\text{]}\\
\\
Remember:\\
- The SQL components must be parseable by DuckDB.\\
- Do not change rows when you generate the VQL.\\
- Always verify your answer carefully before submitting.\\
\end{promptbox}
\end{document}